\documentclass[twoside]{article}

\usepackage[accepted]{aistats2023}
\usepackage[round]{natbib}
\renewcommand{\bibname}{References}
\renewcommand{\bibsection}{\subsubsection*{\bibname}}

\usepackage{amsmath,amsfonts,bm}

\def\eqref#1{equation~\ref{#1}}

\def\1{\bm{1}}

\DeclareMathAlphabet{\mathsfit}{\encodingdefault}{\sfdefault}{m}{sl}
\SetMathAlphabet{\mathsfit}{bold}{\encodingdefault}{\sfdefault}{bx}{n}

\usepackage{graphicx}
\usepackage[utf8]{inputenc} 
\usepackage[T1]{fontenc}    
\usepackage{hyperref}       
\usepackage{url}            
\usepackage{booktabs}       
\usepackage{amsfonts}       
\usepackage{nicefrac}       
\usepackage{microtype}      
\usepackage{xcolor}         
\usepackage{enumitem}
\usepackage{graphicx}
\usepackage{amsmath}
\usepackage[font={small}]{caption}
\usepackage{subcaption}
\usepackage{widetable}
\usepackage{multirow}
\usepackage{multicol}
\usepackage{float} 

\renewcommand{\vec}[1]{\boldsymbol{#1}}

\newcommand{\methodname}{SurvivalGAN}

\begin{document}

%

%
\runningauthor{Alexander Norcliffe, Bogdan Cebere, Fergus Imrie, Pietro Li\`{o}, Mihaela van der Schaar}

\twocolumn[

\aistatstitle{SurvivalGAN: Generating Time-to-Event Data for Survival Analysis}

\aistatsauthor{ Alexander Norcliffe\textsuperscript{*} \And Bogdan Cebere\textsuperscript{*}  \And  Fergus Imrie }

\aistatsaddress{ University of Cambridge\\ \texttt{alin2@cam.ac.uk} 
\And  University of Cambridge \\ \texttt{bcc38@cam.ac.uk}
\And University of California, Los Angeles \\ \texttt{imrie@ucla.edu}} 

\aistatsauthor{Pietro Li\`{o} \And Mihaela van der Schaar}
\aistatsaddress{University of Cambridge \\ \texttt{pl219@cam.ac.uk} 
\And University of Cambridge \\ Alan Turing Institute \\ \texttt{mv472@cam.ac.uk}}
]

\begin{abstract}
Synthetic data is becoming an increasingly promising technology, and successful applications can improve privacy, fairness, and data democratization. While there are many methods for generating synthetic tabular data, the task remains non-trivial and unexplored for specific scenarios. One such scenario is survival data. Here, the key difficulty is censoring: for some instances, we are not aware of the time of event, or if one even occurred. Imbalances in censoring and time horizons cause generative models to experience three new failure modes specific to survival analysis: (1) generating too few at-risk members; (2) generating too many at-risk members; and (3) censoring too early. We formalize these failure modes and provide three new generative metrics to quantify them. Following this, we propose \methodname, a generative model that handles survival data firstly by addressing the imbalance in the censoring and event horizons, and secondly by using a dedicated mechanism for approximating time-to-event/censoring. We evaluate this method via extensive experiments on medical datasets. \methodname{} outperforms multiple baselines at generating survival data, and in particular addresses the failure modes as measured by the new metrics, in addition to improving downstream performance of survival models trained on the synthetic data.
\end{abstract}

\section{INTRODUCTION}

Deep learning has seen incredible success in recent years, yet generally deep models still require large amounts of high-quality data to train well. Data collection is expensive, and often privacy constraints limit how much data can actually be used or shared \citep{de2012data, jain2016big}. Synthetic data generation is a popular solution that aims to create new data that mirrors the statistical properties of the original dataset, tackling the need for privacy and more data in parallel \citep{jordon2022synthetic}. Synthetic data has significant promise, with the potential to improve: (1) fairness \& bias by generating data from underrepresented groups \citep{van2021decaf}; (2) robustness by augmenting an original dataset \citep{perez2017effectiveness}; (3) privacy by not using identifiable data to train a supervised model \citep{zhang_privbayes,jordon2018pate,yoon2020adsgan}, and (4) data democracy by allowing researchers with fewer resources to access inexpensive data \citep{benedetti2020practical, wang2021generating}. As a result, significant attention has been placed on developing generative models \citep{prakash2019structured, tobin2017domain}. Prominent examples can be found across many domains such as images \citep{karras2019style, hinterstoisser2019annotation}, audio \citep{oord2016wavenet}, and medicine \citep{chen2021synthetic, singh2011survival}. One area that remains vastly unexplored is survival data.  

Survival analysis seeks to answer the question: given some measurements at a fixed point in time, how long will it take until a specific event occurs? In engineering for example, given a machine's current condition when do we expect there to be mechanical failure \citep{de2010analysis}, or in finance if a company's stock price is at a certain value how long will it be before they declare bankruptcy. Survival models are incredibly impactful; for instance in medicine \citep{lee1997survival, arsene2007artificial}
they can be used to estimate how long a patient is expected to survive with a given disease
such as Covid-19 \citep{kaso2022survival, lu2021survival, salinas2020survival, ali2022survival}. 
%
They can also be used in clinical trials \citep{singh2011survival} to investigate how long it may be before a death, relapse, or adverse reaction.
%
%
%
Further application areas include economics \citep{danacica2010using, leclere2005preface} and sociology \citep{kent2010predicting,gross2014rate}. Given these models' impact, we wish to generate the highest quality synthetic data for survival analysis.

Naively, generating synthetic data for survival analysis seems straightforward. However, there are two significant obstacles: tabular data and censored data. The tabular setup is notably more complex than image or text data where generative models have typically been applied \citep{xu2019ctgan}. This is due to the mixture of categorical and continuous features and their joint distributions. On top of tabular complexity, in most cases survival data is not completely observed: we may not know when an event occurred -- if at all. This data is said to be ``censored''.
For instance in a drug trial, if the event is being cured of a disease and a subject withdraws from the trial, we will not know when, or even if, they were cured. As a result, the subject is censored at the point of withdrawal.
In spite of missing the time of the event (if indeed an event occurred), censored data still contains information, we know that the event has \emph{not occurred} before the time of censoring. Further, censored data is often more abundant than non-censored, hence, a good generative model will incorporate this data and the imbalance, despite the missing event.
%
%
%
%
%
\paragraph{Contributions.} In this work we present a synthetic  generation framework for efficiently handling censored tabular survival data. Our contributions are three-fold:
\begin{enumerate}[leftmargin=*, noitemsep,topsep=0pt]
    \item We formalize the synthetic data generation problem for survival analysis, identify three failure modes unique to the survival setting, and introduce three metrics to quantitatively evaluate these failures and provide a clearer understanding of the synthetic data's utility (Sections \ref{section:formulation} \& \ref{sec:metrics}).
    \item  We propose \methodname, a method that is able to efficiently incorporate censored data, tabular data, and censoring \& time-horizon imbalance to generate synthetic data to train survival models (Section \ref{section:solution}). 
    \item We investigate \methodname{} via extensive experiments on five medical datasets. We demonstrate its ability to generate high-quality synthetic data relative to five robust benchmarks, as measured by established generative metrics, the new survival specific metrics, and downstream model performance (Section \ref{section:experiments}).
\end{enumerate}

\section{RELATED WORK} 

\paragraph{Generative Models.} Generative models come in various flavors. Classically, Bayesian networks represent a high dimensional distribution with a directed acyclic graph to compactly give the dependency structure \citep{niedermayer2008introduction}. These can often be slow to sample from, requiring methods such as Markov Chain Monte-Carlo. More recent generative models include Variational Auto-Encoders \citep{kingma2013auto, vahdat2020nvae}, Generative Adversarial Networks \citep{goodfellow2014generative} and Normalizing Flows \citep{rezende2015variational, kingma2018glow}. Exact training techniques and architectures differ, but typically modern models use deep networks to learn a mapping from an easy-to-sample latent space (such as Gaussian noise) to the data distribution, leading to fast sampling. The exception is diffusion-based models \citep{song2019generative, ho2020denoising, ramesh2022hierarchical} which learn a reverse diffusion process in observation space to take noisy points to regions of higher probability via a series of (possibly many) function evaluations. 

In their standard implementations, these methods are not well suited to tabular data, due to the mixture of continuous and categorical variables. 
Variants have been proposed that are designed to work in the tabular domain. Tabular GAN \citep{xu2018synthesizing} works on tabular data and CTGAN \citep{xu2019ctgan} is a GAN designed to work specifically with tabular data that mixes categorical and continuous variables. This is achieved with a tabular encoding and training by sampling. TabFairGAN \citep{rajabi2022tabfairgan} extends this to include a fairness constraint generating accurate and fair data. Tabular variants of other models also exist, such as TVAE \citep{xu2019ctgan}, RTVAE \citep{akrami2022robust}, and GOGGLE \citep{liu2023goggle}, which adapt VAEs, while \citet{vahdat2021score} adapt score-based models to generate tabular data by running the diffusion in latent space. These principled methods are well suited to tabular data, however, survival data contains the added difficulty of censored data which these models are not able to handle.


\paragraph{Survival Data.} 
Few models exist to generate synthetic survival data. Existing survival models learn to sample from the conditional distribution of event time given the initial state (known as the covariates), for example \citet{bender2005generating} and \citet{austin2012generating} present statistical models that transform samples from a uniform distribution to survival times by inverting the cumulative hazard function, conditioned on the covariates. Sampling from $p(t|\vec x)$ is a probabilistic survival model and does not generate the covariates, which we desire to make a full synthetic dataset.
More modern techniques incorporate deep learning in the generative process. \citet{ranganath2016deep} use deep exponential families to generate survival data, limiting the flexibility of the learnt distribution. \citet{miscouridou2018deep} and \citet{zhou2022deep} relax this assumption but still generate survival times and censoring statuses conditioned on covariates, rather than generating covariates and times. None of these models are able to generate survival data that considers censoring and still generates covariates with high fidelity.

\section{PROBLEM FORMULATION AND BACKGROUND}
\label{section:formulation}

\subsection{Generating Synthetic Survival Data}
\label{subsection:survivalformulation}

Synthetic data generation is concerned with creating new data points that resemble but are not identical to real data. Given a dataset of $N$ training observations, $\mathcal{D} = \{ \vec z_i \}_{i=1}^{N}$ drawn from an unknown distribution $\vec z_{i} \sim p_{\vec z}$, the task is to generate $M$ new observations $\mathcal{D}^{\text{Syn}} = \{\vec z_j^{\text{Syn}}\}_{j=1}^{M}$ that appear to have been sampled from the same distribution. 
Instances from survival data are of the form $\vec z_i = (\vec x_i, t_i, E_i)$. Here, $\vec x \in \mathcal{X}$ are $m$-dimensional tabular covariates giving a subject's state at an initial time, containing continuous and categorical features. $t_i \in \mathcal{T}$ is the time of a given event, usually the initial time is 0 such that $\mathcal{T} = \mathbb{R}_{\geq 0}$ . Finally, $E \in \mathcal{E}$ is the event indicator, typically $\mathcal{E} = \{0, 1\}$, where $E=0$ means the individual is censored at $t$, and $E=1$ means the event of interest occurred at $t$. In this exposition we consider a single event, but this can be naturally extended to competing events by extending $\mathcal{E}$ to include new discrete events.

\subsection{Survival Analysis}

Here we give a brief summary of classical survival analysis\footnote{Predicting the time of event given a covariate, not generating survival data itself.} relevant to this work. This is meant as a way to introduce key terms from survival analysis that we will use throughout the paper; for a thorough introduction see \citet{jenkins2005survival} and \citet{machin2006survival}. 

\textbf{Survival function.} 
We can conduct survival analysis with the probability density function $p(t|\vec x)$, giving the likelihood that the event of interest happens at time $t$ given covariate $\vec x$. With this, the \textbf{survival function} is defined as the probability that the event \emph{has not} occurred by time $t$,
%
\begin{equation}
\label{eq:survival}
  S(t|\vec x)=\int_{t}^{\infty}p(t'|\vec x)dt'.  
\end{equation}
Equation (\ref{eq:survival}) gives us the proportion of the subjects with covariate $\vec x$ that have survived up to point $t$. When the initial time is zero, the event cannot happen before $t=0$, so $S(0|\vec x)=1$, and $p(t|\vec x)$ is a valid probability distribution (non-negative) so $S(t|\vec x)$ is a decreasing function. 
%

\textbf{Time-to-event approximation.} This is the task of approximating the expected lifetime for any $\vec x$. By definition this is given by $\mu(\vec x) = \int_{0}^{\infty}t'p(t'|\vec x)dt'$. Via integration by parts, this is equal to the area under the survival curve $\mu(\vec x) = \int_0^\infty S(t|\vec x) dt$. 
The above properties can be calculated at a population level by marginalizing out $\vec x$. Survival models usually fall into one of these two categories, i.e. (1) estimate the survival function, or (2) estimate the time-to-event.
The first category is well-studied, with solutions ranging from linear models \citep{cox1972regression} to random forests \citep{ishwaran2008random} to gradient boosting \citep{barnwal2022survival}. The second category can be derived from the survival probabilities, but has also been studied independently using neural networks \citep{lee2018deephit,chapfuwa2018adversarial}. 
Finally, it is possible to combine approaches from both categories via ensembling methods \citep{imrie2022autoprognosis}.

\subsection{Challenges of Generating Survival Data}
\label{subsection:challenges}

\paragraph{Tabular Data.} Handling tabular data is non-trivial for most generative models. Some of the reasons are the mixed feature types (categorical or continuous), non-Gaussian distribution of the features, or highly imbalanced categorical features \citep{xu2019ctgan}. 
%
\paragraph{Time and Censoring Failure Modes.}
We are attempting to generate data from the full distribution $p(\vec x, t, E)$; generating covariates $\vec x$ is subject to the traditional failure modes of generative modeling. However, modeling the marginal distribution of the time and event pair $p(t, E)$ is specific to survival analysis and introduces three new failure modes, driven by two key dataset imbalances:
\begin{enumerate}[leftmargin=*, noitemsep,topsep=0pt]
    \item \textbf{Censoring imbalance:} There is often an imbalance in the amount of censoring. By not incorporating this balance, a model could generate unusable data, for example consisting of too many censored records. This makes a model \textbf{over-optimistic} because the event in question occurs less frequently than expected and the model predicts too high an expected lifetime; this is the first failure mode. On the other hand, by not generating enough censored data, the predicted lifetime could become too low, resulting in the model being \textbf{over-pessimistic}; this is the second failure mode.
    \item \textbf{Time imbalance:} It is also possible for datasets to have a time-horizons imbalance. Where the majority of events are early, $t$ is early regardless of $E$. A model is at risk of learning this imbalance and making events too early. We don't want to focus only on short-term examples we want to have a broad view of the timeline. Such a model is \textbf{short-sighted}; this is the third failure mode. 
\end{enumerate}

\begin{figure*}[ht]
     \centering
     \begin{subfigure}[b]{0.32\textwidth}
         \centering
         \includegraphics[width=1.05\textwidth]{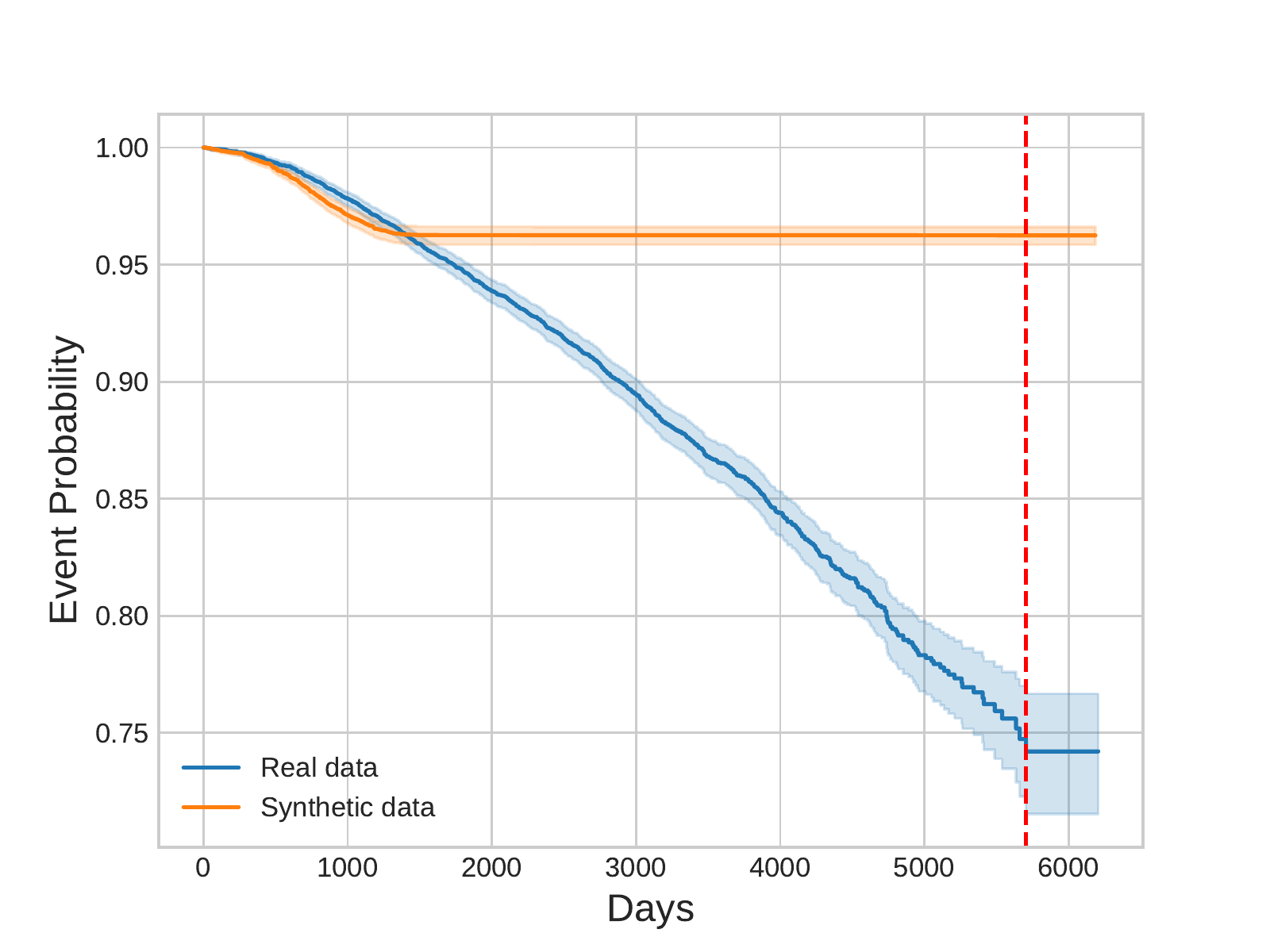}
         \caption{Over-optimism. The area under the synthetic curve is larger than the true curve, the model is over-optimistic.}
     \end{subfigure}
     \hfill
     \begin{subfigure}[b]{0.32\textwidth}
         \centering
         \includegraphics[width=1.05\textwidth]{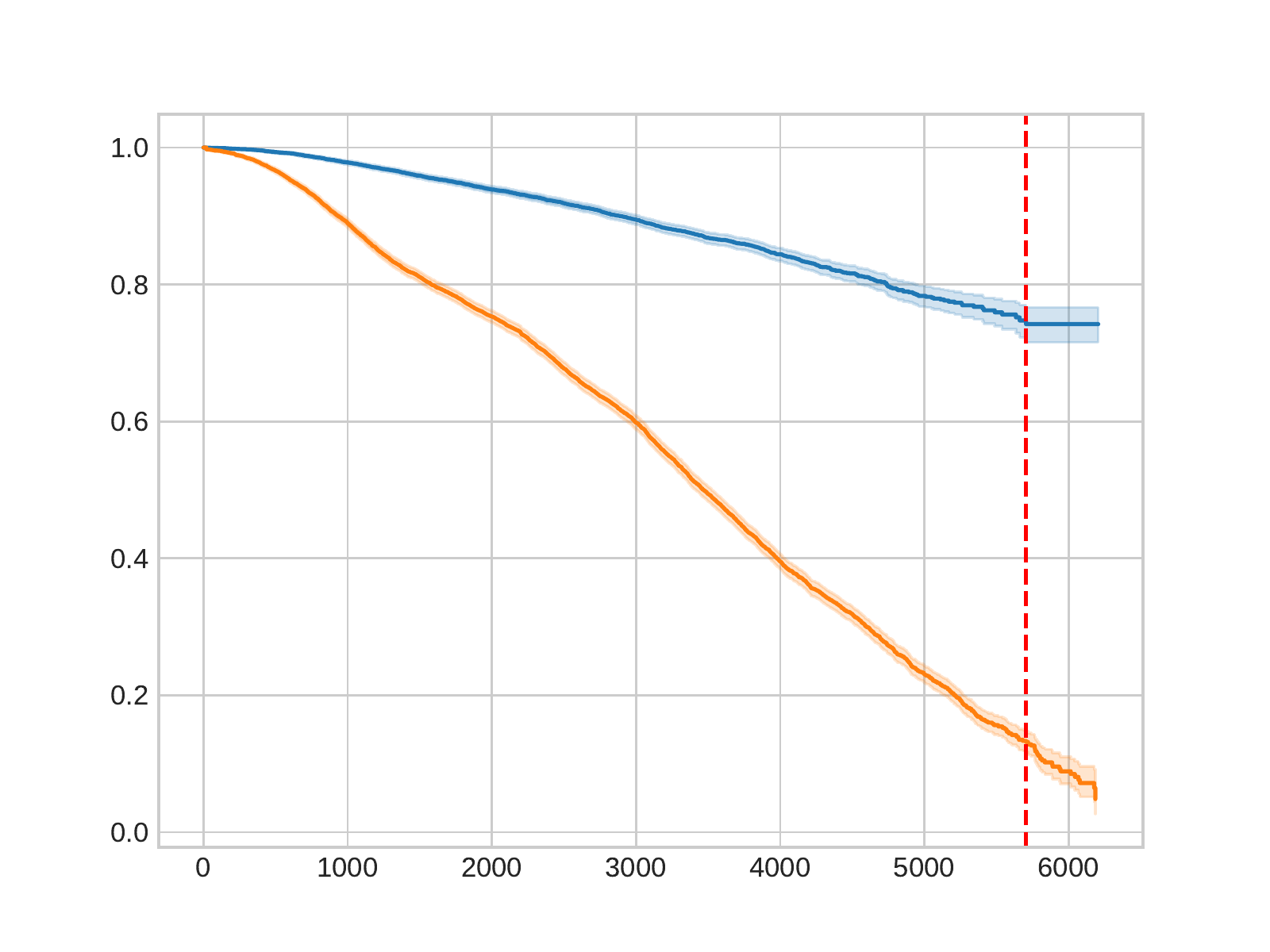}
         \caption{Over-pessimism. The area under the synthetic curve is lower than the true curve, the model is over-pessimistic.}
     \end{subfigure}
     \hfill
     \begin{subfigure}[b]{0.32\textwidth}
         \centering
         \includegraphics[width=1.05\textwidth]{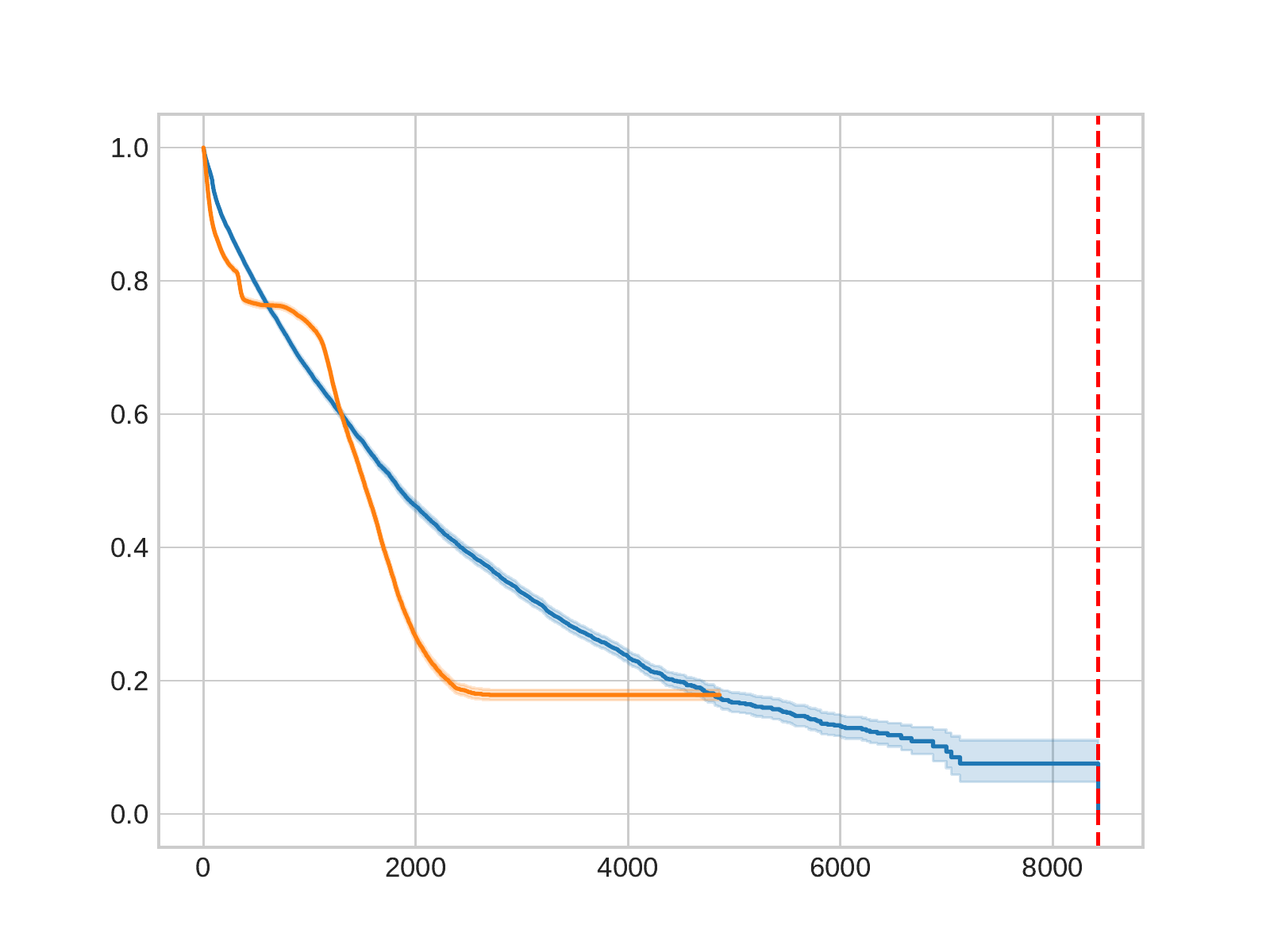}
         \caption{Short-sightedness. The synthetic curve ends noticeably earlier than the true curve, the model is short-sighted.}
     \end{subfigure}
        \caption{The three possible failure modes specific to survival data, illustrated with Kaplan-Meier plots.}
        \label{fig:failure_modes_example}
\end{figure*}

We illustrate these failure modes in Figure \ref{fig:failure_modes_example}. These show the Kaplan-Meier (KM) curves \citep{kaplan1958nonparametric}, which plot an approximation of the survival function at a given time. For a finite sample size, the KM curves show the proportion of the subjects that have made it to time $t$ without the event happening. We assume the sample size is large enough, such that the curve approaches the true survival function. To quantify these failure modes we introduce three metrics in the next section.

\section{ASSESSING SYNTHETIC SURVIVAL DATA}
\label{sec:metrics}

Various metrics exist to evaluate the similarity between synthetic data and real data. For example, Maximum Mean Discrepancy \citep{sutherland2017generative}, Inception Score \citep{salimans2016improved}, and Fr\'{e}chet Inception Distance \citep{heusel2017gans} are used to evaluate sample quality. However, representing the performance of a synthetic dataset with one metric is prone to over-simplifying the situation. In \citet{alaa2022faithful}, three further model-independent metrics are proposed to overcome this, $\alpha$-Precision $\beta$-Recall and Authenticity. These are designed to evaluate sample quality, diversity, and similarity to real data, respectively. Importantly, none of these metrics are designed for survival data and are unable to capture the failure modes that originate from censoring imbalance and time imbalance. 

We introduce three new metrics specific to evaluating synthetic survival data, targeting the failure modes in modeling the marginal distribution of $p(t, E)$. We have seen that as well as traditional generative failure modes (e.g. mode collapse or low sample quality), survival data presents three new ones: over-optimism, over-pessimism, and short-sightedness. We quantify these phenomena by looking at the differences in survival probabilities. Survival curves have been used to quantify the performance of predictive survival models but have not been used to evaluate synthetic survival data.


\paragraph{Optimism.} The total area under the Kaplan-Meier plot gives us the expected lifetime of the population $\mu = \int_{0}^{\infty}S(t)dt$. We call the model over-optimistic if $\mu_{\text{Syn}} > \mu_{\text{Real}}$, and over-pessimistic for the reverse. Therefore we are interested in the quantity $\mu_{\text{Syn}} - \mu_{\text{Real}}$. The KM plots are only available in a finite interval, after which censoring is present. This can make the expected lifetime diverge if the survival function has not reached zero by this point. Instead, we consider the plots up to the latest available time $T$. If the synthetic KM plot ends before the real one (due to short-sightedness), it is extrapolated assuming a constant rate of events. This gives the definition of optimism
\begin{equation}
    \text{Optimism} = \frac{1}{T}\int_{0}^{T}
    \big(
    S_{\text{Syn}}(t) - S_{\text{Real}}(t)
    \big)
    dt.
\end{equation}
This is the mean difference between the two Kaplan-Meier plots, and can also be viewed as a scaled difference in their areas. This metric takes values between -1 and 1, with 0 predicting the exact same expected lifetime, not over-optimistic or over-pessimistic. Positive values represent a synthetic expected lifetime higher than the true one, making the data over-optimistic and vice versa.

\paragraph{Short-Sightedness.} Models trained on synthetic data may not be able to predict past a certain time horizon, meaning that the synthetic data is censored from that point. 
To quantify this using the KM plots, we consider the two end times $T_{\text{Syn}}$ and $T_{\text{Real}}$ and take their relative difference:
\begin{equation}
\label{eq:short_sightedness}
    \text{Short-Sightedness} = \frac{T_{\text{Real}} - T_{\text{Syn}}}{T_{\text{Real}}}.
\end{equation}
This quantifies the relative amount that the time horizons in the generated data stop before the real data. This metric takes values between 0 and 1, with 0 giving no short-sightedness and 1 giving full short-sightedness.

This metric can be generalized to also measure long-sightedness, where the predicted times are larger than the true times. Instead of dividing by $T_{\text{Real}}$, we divide by $\max(T_{\text{Real}}, T_{\text{Syn}})$, giving a value between -1 and 1, 0 being perfect, 1 being maximally short-sighted and -1 being maximally long-sighted. We observed that this does not happen in practice and therefore use the simpler definition in Equation (\ref{eq:short_sightedness}).

\paragraph{Kaplan-Meier Divergence.} It is possible to have scores of zero for both optimism and short-sightedness but still have non-matching Kaplan-Meier curves. Therefore we finally include the mean absolute difference between the curves, which we call the Kaplan-Meier divergence (KM Divergence),
\begin{equation}
    \text{KM Divergence} = \frac{1}{T}\int_{0}^{T}
    \big|
    S_{\text{Syn}}(t)-S_{\text{Real}}(t)
    \big|
    dt.
\end{equation}
This will be between 0 and 1 (because $S(t)$ is always between 0 and 1), with 0 when the curves match perfectly, and 1 when they have the maximum difference possible at all times. In Appendix \ref{app:metrics}, we demonstrate the need for all three metrics, showing they target different ways the KM curves can differ; with optimism and short-sightedness having an interpretable meaning. We also show that it is possible to bound the optimism by the total-variation divergence of the underlying probability density functions.

\paragraph{Other Metrics.} Besides measuring the quality of $p(t, E)$ using these new metrics, we can measure the quality of the covariate marginal distribution $p(\vec x)$ using a standard generative metric of choice. To measure the quality of the full distribution $p(\vec x, t, E)$, we evaluate the downstream performance of models trained with synthetic data compared to those trained with real data.

\begin{figure*}[ht!]
     \centering
     \includegraphics[width=\textwidth]{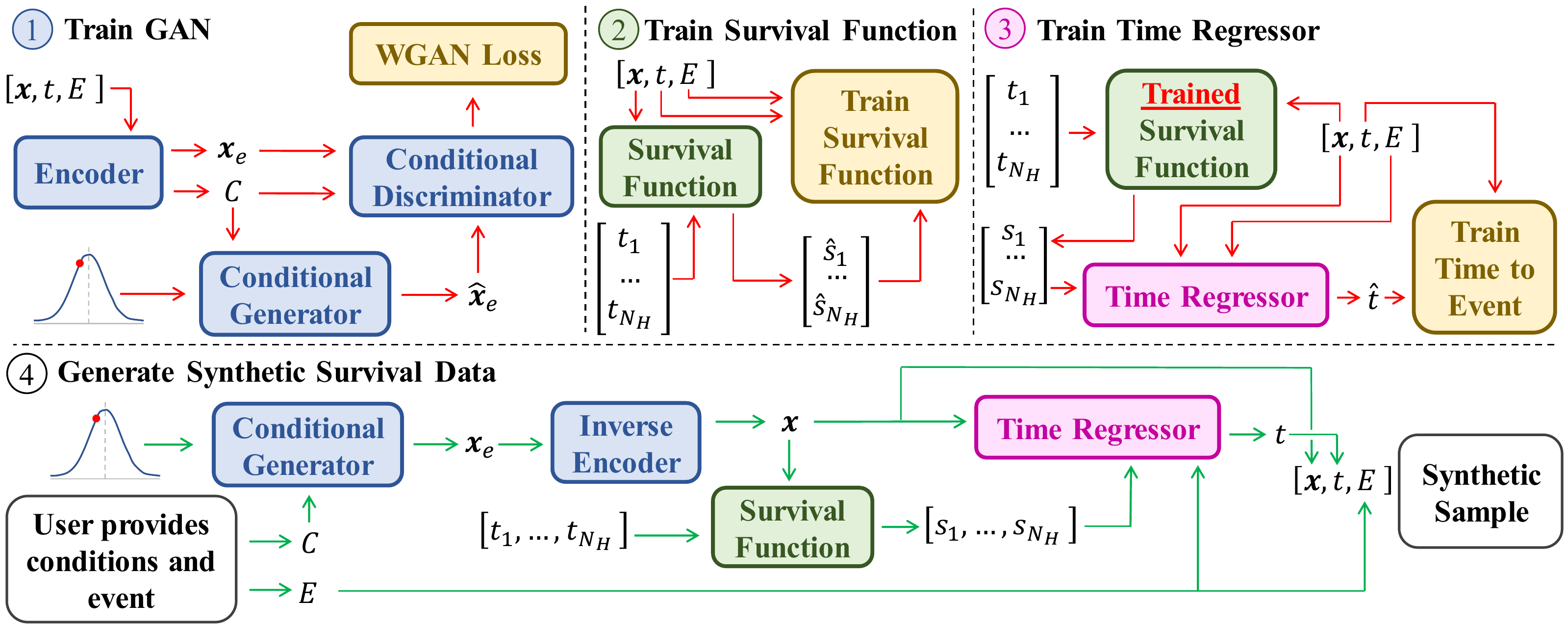}
     \caption{Block diagram of \methodname{}'s components, their interaction, and the generation process. Both the specific survival function and time regressor may be chosen, hence the generic training for those components. Arrows pointing to the parentheses around vectors imply the whole vector is the input/output of a function. Arrows pointing to specific vector components only use those.}
     \label{fig:survival_gan_schematic}
\end{figure*}

\section{OUR MODEL - SURVIVALGAN}
\label{section:solution}

Below we describe our solution -- \methodname{} -- in depth. A block diagram is given in Figure \ref{fig:survival_gan_schematic}. Briefly, to generate synthetic data, a condition vector $C$ and event $E$ are given by the user. The condition is a one-hot vector with both interpretable features (sex for example) and latent encodings (see full description below) and the event indicates censoring or true event ($0$ or $1$). These can either be sampled according to the training data frequencies ($C, E \sim p_{C, E}$) or by manually selecting them. A conditional GAN is used to generate a covariate $\vec x \sim p_{\vec x | C, E}$. Finally, a survival function and time-to-event/censoring regressor are used together to generate the time $t \sim p_{t| \vec x, E}$. One key insight is that the method assumes that censoring occurs independently and at random, and given some new covariates and a censoring status, a separate model determines the time of event/censoring. This allows us to follow a similar censoring ratio as the training dataset, without risking censoring all the synthetic instances.

\subsection{Model and Training}
\label{subsection:encoders}

\paragraph{Conditional GAN.} The conditional GAN (part 1 of Figure \ref{fig:survival_gan_schematic}), allows us to sample $\vec x \sim p_{\vec x|C, E}$, where $C$ is a user determined condition that the generator takes as input. We start by training a tabular encoder. This is critical as: (1) it allows us to handle censoring and time imbalance (as well as imbalance in the covariates) during training and generation and (2) it enables us to handle continuous and categorical variables; these were the key difficulties specific to survival data as described in Section \ref{subsection:challenges}. We follow the approach by \citet{xu2019ctgan}. For each continuous feature, a Gaussian Mixture Model (GMM) \citep{reynolds2009gaussian} with $N_{\text{C}}$ components is trained. The tabular encoder for that feature is then given by $\text{Encoder}: \mathbb{R} \xrightarrow{} \{0, 1\}^{N_{\text{C}}} \times \mathbb{R}$, where the first part of the output is a one-hot encoding telling us which component of the GMM the feature belongs to, and the second part is the number of standard deviations away from that component's mean $(x-\mu)/\sigma$. Note that the inverse of this encoding is trivial and does not need to be learnt since we include the one hot encoding of which mode the feature is in, and we know $\mu$ and $\sigma$ for that mode. For categorical features, the encoding is simply a one-hot encoding of the feature. Following this we also define a class encoder, $\text{ClassEncoder} : \mathbb{R} \xrightarrow{} \{0, 1\}^{N_{\text{C}}}$, which simply takes the one-hot vector from the tabular encoding, saying which GMM component the feature belongs to without the location within that component. This allows us to represent a condition, $C$, for the generator. For extensive information on the tabular encoding see \citet{reynolds2009gaussian}.       

Once the tabular encoder has been trained, we train the GAN, which consists of generator $G_{\theta_g}$ and discriminator $D_{\theta_d}$. For a given sample $(\vec x, t, E)$ from the training dataset, the covariate is encoded first, $\vec x_{e} = \text{Encoder}(\vec x)$. Following this, the condition is encoded from the input $C = \text{ClassEncoder}(\vec x, t, E)$. We then sample $\epsilon \sim \mathcal{U}_{[0, 1]}$ and $\vec z \sim \mathcal{N}(0, I)$ and train with the Wasserstein GAN losses with gradient penalty:
%
\begin{align*}
\tilde{\vec x}_{e} &= \epsilon  \vec x_{e} + (1 - \epsilon) G_{\theta_g}(C, \vec z)
\\
L_{G} &= -D_{\theta_d}(C, G_{\theta_g}(C, \vec z))
\\
L_{D} &= D_{\theta_d}(C, G_{\theta_g}(C, \vec z)) 
-
D_{\theta_d}(C, \vec x_{e})
\\
& \qquad \qquad +\lambda(\lVert \nabla_{\tilde{\vec x}_{e}}D_{\theta_d}(C, \tilde{\vec x}_{e})\rVert_{2} - 1)^2,
\end{align*}
where $\lambda$ is the gradient penalty. We found that using the Wasserstein GAN with gradient penalty and the tabular encoder made training more stable.

\paragraph{Survival Function.} The survival function (part 2 of Figure \ref{fig:survival_gan_schematic}), is used with a time regressor (see next model component) to sample from $p_{t|\vec x, E}$. The survival function $S: \mathcal{X} \times \mathcal{T} \xrightarrow{} [0,1]$ is a decreasing function predicting the survival probability at a given time horizon for given covariates $\vec x$. This is used in the generation process to create a set of survival probabilities at $N_{\text{H}}$ time horizons $\{S(\vec x, t_i)\}_{i=1}^{N_\text{H}}$. In practice, we use 100 evenly spread times between the minimum and maximum horizons in the training data. Any survival model can be used in this setup, thus the training is general. We use DeepHit \citep{lee2018deephit} because, beyond strong predictive performance, it can be extended to competing risks, making it a flexible solution for future extensions. DeepHit uses a custom loss function consisting of two terms, one for log-likelihood of the joint distribution of $t$ and $E$, and one incorporating cause-specific losses. We refer the reader to \citet{lee2018deephit} for more detail.

\paragraph{Time Regressor.} The time regressor (part 3 of Figure \ref{fig:survival_gan_schematic}) is trained after the survival function as it relies on using a trained survival model. It is used with the survival function to sample from $p_{t|\vec x, E}$. The time regressor is a function $T: \mathcal{X} \times [0, 1]^{N_\text{H}} \times \mathcal{E} \xrightarrow{} \mathcal{T}$, which takes as input a covariate, event type and $N_{\text{H}}$ outputs of the survival function. The function then predicts the time that the event or censoring happens such that
$T(\vec x, S(\vec x, t_1), S(\vec x, t_2),...,S(\vec x, t_{N_{\text{H}}}), 0)$ gives the time to \emph{censoring} for a given covariate, and 
$T(\vec x, S(\vec x, t_1), S(\vec x, t_2),...,S(\vec x, t_{N_{\text{H}}}), 1)$ is the time to \emph{event} for a given covariate.
%
%
Any time regressor may be used, hence the training is general. We use XGBoost \citep{chen2016}, with the mean squared error of $\log(t)$ as the loss.

\subsection{Generation}

Generation from the model is shown in part 4 of Figure \ref{fig:survival_gan_schematic}. First, a one-hot condition vector $C$ and type of event $E$ are provided by the user. The user can choose these, allowing us to sample underrepresented groups within the covariates, and target specific edge cases when creating a synthetic dataset if desired. Alternatively, and the method we opt for in our empirical evaluation, one can sample conditions from a categorical distribution constructed from training data frequencies, which we call the \textbf{Imbalanced Sampler}. The generator uses this condition vector to produce an encoded covariate $\vec x_{e} = G_{\theta_g}(C, \vec z)$, where $\vec z \sim \mathcal{N}(0, I)$. This is deterministically inverted to produce the covariate $\vec x = \text{Encoder}^{-1}(\vec x_{e})$. Finally, this is used in the survival function with predefined time horizons and the time regressor to generate the time of event/censoring $t=T(\vec x, S(\vec x, t_1), S(\vec x, t_2),...,S(\vec x, t_{N_{\text{H}}}), E)$.

\paragraph{Necessity of Components.} To demonstrate that all components of \methodname{} are necessary, we carry out an ablation study in Section \ref{subsection:experiments_gains} individually testing each module. Table \ref{table:experiments_gain_source} demonstrates that all are crucial for the quality of the synthetic survival data.

\begin{table*}[h]\scriptsize
\caption{The mean and standard deviation of the covariate and time-censoring metrics. The best values are in \textbf{bold}, for short-sightedness extreme failure cases are \underline{underlined}. $*$: Not evaluated.}
\label{table:experiments_data_fidelity}
\begin{widetable}{\textwidth}{c|c|c|c|c|c|c}
\toprule
Metric & Method & AIDS  & CUTRACT & PHEART & SEER & METABRIC  \\
\midrule
\multirow{6}*{
    \begin{tabular}{c}
        Jensen-Shannon Distance \\
        \tiny{(Lower is Better)}
    \end{tabular}
}  
		
& SurvivalGAN & 
    \multicolumn{1}{r|}{\textbf{0.012$\pm$0.02}} &
    \multicolumn{1}{r|}{0.024$\pm$0.01} & 
    \multicolumn{1}{r|}{0.013$\pm$0.01} & 
    \multicolumn{1}{r|}{0.022$\pm$0.01} &
    \multicolumn{1}{r}{0.015$\pm$0.01} \\
& PrivBayes & 
    \multicolumn{1}{r|}{0.028$\pm$0.01} & 
    \multicolumn{1}{r|}{0.020$\pm$0.01} & 
    \multicolumn{1}{r|}{\textbf{0.008$\pm$0.01}} &
    \multicolumn{1}{r|}{0.022$\pm$0.01} &
    \multicolumn{1}{r}{0.043$\pm$0.01} \\
& ADS-GAN & 
    \multicolumn{1}{r|}{0.052$\pm$0.01} & 
    \multicolumn{1}{r|}{0.054$\pm$0.02} & 
    \multicolumn{1}{r|}{0.039$\pm$0.02} & 
    \multicolumn{1}{r|}{0.036$\pm$0.01} &
    \multicolumn{1}{r}{0.041$\pm$0.01} \\
& CTGAN & 
    \multicolumn{1}{r|}{0.031$\pm$0.01} & 
    \multicolumn{1}{r|}{\textbf{0.011$\pm$0.01}} & 
    \multicolumn{1}{r|}{0.015$\pm$0.01} & 
    \multicolumn{1}{r|}{\textbf{0.008$\pm$0.01}} &
    \multicolumn{1}{r}{0.015$\pm$0.01} \\
& TVAE & 
    \multicolumn{1}{r|}{0.054$\pm$0.01} & 
    \multicolumn{1}{r|}{0.040$\pm$0.01} & 
    \multicolumn{1}{r|}{0.017$\pm$0.01} & 
    \multicolumn{1}{r|}{0.034$\pm$0.01} &
    \multicolumn{1}{r}{0.013$\pm$0.01} \\
& NFlows & 
    \multicolumn{1}{r|}{0.048$\pm$0.01	} & 
    \multicolumn{1}{r|}{0.038$\pm$0.01} & 
    \multicolumn{1}{r|}{0.044$\pm$0.01	} & 
    \multicolumn{1}{r|}{0.030$\pm$0.01} &
    \multicolumn{1}{r}{\textbf{0.007$\pm$0.01}} \\
    
\midrule
\multirow{6}*{
    \begin{tabular}{c}
        Wasserstein Distance \\
        \tiny{(Lower is Better)}
    \end{tabular}
}  
		
& SurvivalGAN & 
    \multicolumn{1}{r|}{\textbf{0.153$\pm$0.02}} & 
    \multicolumn{1}{r|}{0.228$\pm$0.11} & 
    \multicolumn{1}{r|}{\textbf{0.441$\pm$0.12}} & 
    \multicolumn{1}{r|}{0.420$\pm$0.29} &
    \multicolumn{1}{r}{\textbf{5.861$\pm$3.01}} \\
& PrivBayes & 
    \multicolumn{1}{r|}{1.080$\pm$0.27	} & 
    \multicolumn{1}{r|}{0.101$\pm$0.01} & 
    \multicolumn{1}{r|}{0.828$\pm$0.02	} & 
    \multicolumn{1}{r|}{0.146$\pm$0.01	} &
    \multicolumn{1}{r}{15.718$\pm$4.39} \\
& ADS-GAN & 
    \multicolumn{1}{r|}{1.694$\pm$0.51} & 
    \multicolumn{1}{r|}{2.393$\pm$0.21} & 
    \multicolumn{1}{r|}{2.207$\pm$0.55} & 
    \multicolumn{1}{r|}{2.108$\pm$0.01} &
    \multicolumn{1}{r}{17.806$\pm$0.28} \\
& CTGAN & 
    \multicolumn{1}{r|}{0.851$\pm$0.07} & 
    \multicolumn{1}{r|}{\textbf{0.032$\pm$0.01}} & 
    \multicolumn{1}{r|}{0.621$\pm$0.05	} & 
    \multicolumn{1}{r|}{\textbf{0.019$\pm$0.01}} &
    \multicolumn{1}{r}{9.343$\pm$3.81} \\
& TVAE & 
    \multicolumn{1}{r|}{1.967$\pm$0.05} & 
    \multicolumn{1}{r|}{1.430$\pm$0.20} & 
    \multicolumn{1}{r|}{0.892$\pm$0.30} & 
    \multicolumn{1}{r|}{1.902$\pm$0.10} &
    \multicolumn{1}{r}{9.421$\pm$2.85} \\
& NFlows & 
    \multicolumn{1}{r|}{1.701$\pm$0.19} & 
    \multicolumn{1}{r|}{0.325$\pm$0.05} & 
    \multicolumn{1}{r|}{2.086$\pm$0.03} & 
    \multicolumn{1}{r|}{0.169$\pm$0.04} &
    \multicolumn{1}{r}{14.649$\pm$0.76} \\    

\midrule
\midrule
\multirow{6}*{
    \begin{tabular}{c}
        Optimism \\
        \tiny{(Closer to Zero is Better)}
    \end{tabular}
}  
		
& SurvivalGAN & 
    \multicolumn{1}{r|}{\textbf{-0.006$\pm$0.01}} & 
    \multicolumn{1}{r|}{\textbf{-0.012$\pm$0.03}} & 
    \multicolumn{1}{r|}{\textbf{-0.036$\pm$0.05}} & 
    \multicolumn{1}{r|}{-0.079$\pm$0.02} &
    \multicolumn{1}{r}{-0.113$\pm$0.01} 
    \\
& PrivBayes & 
    \multicolumn{1}{r|}{\textbf{0.006$\pm$0.02}} & 
    \multicolumn{1}{r|}{-0.017$\pm$0.01} & 
    \multicolumn{1}{r|}{0.129$\pm$0.01} & 
    \multicolumn{1}{r|}{\textbf{-0.001$\pm$0.01}}&
    \multicolumn{1}{r}{0.384$\pm$0.01}  \\
& ADS-GAN & 
    \multicolumn{1}{r|}{0.038$\pm$0.04} & 
    \multicolumn{1}{r|}{0.113$\pm$0.01} & 
    \multicolumn{1}{r|}{0.096$\pm$0.21} &  
    \multicolumn{1}{r|}{0.026$\pm$0.01}&
    \multicolumn{1}{r}{0.290$\pm$0.03}  \\
& CTGAN & 
    \multicolumn{1}{r|}{-0.061$\pm$0.03} & 
    \multicolumn{1}{r|}{-0.334$\pm$0.02} & 
    \multicolumn{1}{r|}{-0.099$\pm$0.02} & 
    \multicolumn{1}{r|}{-0.386$\pm$0.04}&
    \multicolumn{1}{r}{0.183$\pm$0.08} 	\\
& TVAE & 
    \multicolumn{1}{r|}{0.064$\pm$0.01} & 
    \multicolumn{1}{r|}{0.079$\pm$0.01} & 
    \multicolumn{1}{r|}{-0.115$\pm$0.03} & 
    \multicolumn{1}{r|}{0.024$\pm$0.01}&
    \multicolumn{1}{r}{0.140$\pm$0.03}  \\
& NFlows & 
    \multicolumn{1}{r|}{-0.183$\pm$0.05} & 
    \multicolumn{1}{r|}{-0.236$\pm$0.06} & 
    \multicolumn{1}{r|}{0.091$\pm$0.02} & 
    \multicolumn{1}{r|}{-0.398$\pm$0.10}&
    \multicolumn{1}{r}{\textbf{0.003$\pm$0.01}}  \\
\midrule
\multirow{6}{*}{
    \begin{tabular}{c}
        KM Divergence \\
        \tiny{(Lower is Better)}
    \end{tabular}}   
& SurvivalGAN & 
    \textbf{0.011$\pm$0.01} & 
    \textbf{0.031$\pm$0.02} & 
    \textbf{0.079$\pm$0.01} & 
    0.080$\pm$0.02 &
    0.121$\pm$0.01
    \\
& PrivBayes & 
    0.035$\pm$0.01 & 
    0.034$\pm$0.01 & 
    0.134$\pm$0.01 & 
    \textbf{0.007$\pm$0.01} &
    0.383$\pm$0.01
    \\
& ADS-GAN & 
    0.054$\pm$0.02 & 
    0.113$\pm$0.01 & 
    0.167$\pm$0.15 & 
    0.026$\pm$0.01 &
    0.308$\pm$0.03
    \\
& CTGAN & 
    0.073$\pm$0.02 & 
    0.334$\pm$0.02 & 
    0.099$\pm$0.02 & 
    0.386$\pm$0.04 &
    0.133$\pm$0.04
    \\
& TVAE & 
    0.064$\pm$0.01 & 
    0.083$\pm$0.01 & 
    0.115$\pm$0.03 & 
    0.024$\pm$0.01 &
    0.168$\pm$0.03
    \\
& NFlows & 
    0.185$\pm$0.05 & 
    0.238$\pm$0.06 &
    0.100$\pm$0.01 & 
    0.392$\pm$0.10 &
    \textbf{0.054$\pm$0.01}
    \\
\midrule
\multirow{6}{*}{
    \begin{tabular}{c}
        Short-sightedness \\
        \tiny{(Closer to Zero is Better)}
    \end{tabular}}
& SurvivalGAN & 
    0.007$\pm$0.01 & 
    0.046$\pm$0.05 & 
    0.127$\pm$0.15 & 
    0.010$\pm$0.04 &
    \textbf{0.027$\pm$0.02} 
    \\ 
& PrivBayes & 
    0.013$\pm$0.01 & 
    0.004$\pm$0.01 &
    0.135$\pm$0.01 &
    \textbf{0.000$\pm$0.00} &
    $*$ 
    \\ 
& ADS-GAN & 
    0.074$\pm$0.02 & 
    0.137$\pm$0.10 & 
    \underline{0.479$\pm$0.05} & 
    0.080$\pm$0.05 &
    $*$ 
    \\
& CTGAN  & 
    \textbf{0.000$\pm$0.00} &
    \textbf{0.001$\pm$0.01} & 
    \underline{0.438$\pm$0.01} & 
    \textbf{0.000$\pm$0.00} &
    0.131$\pm$0.03 
    \\
& TVAE  & 
    \textbf{0.000$\pm$0.00} & 
    0.003$\pm$0.01 & 
    \underline{0.410$\pm$0.01} & 
    \textbf{0.000$\pm$0.00} &
    0.147$\pm$0.03 
    \\
& NFlows  & 
    0.001$\pm$0.01 &
    0.002$\pm$0.01 &
    \textbf{0.106$\pm$0.09} &
    0.0001$\pm$0.01 &
    $*$ 
    \\
\bottomrule
\end{widetable}
\end{table*}

\section{EXPERIMENTS}
\label{section:experiments}

To assess the quality of \methodname{}, we evaluate the following  aspects:
\begin{enumerate}[leftmargin=*, noitemsep,topsep=0pt]
    \item \textbf{Quality of Marginals:} 
    Section \ref{subsection:experiments_diversity} analyses how closely the distribution of synthetic samples matches the original from two perspectives: (1) the marginal distribution of covariates $p(\vec x)$. This is done using Jensen-Shannon distance and Wasserstein distance between real and synthetic covariates. We do not necessarily expect \methodname{} to perform better than the baselines here, but we check it does not perform worse.
    (2) the censoring and temporal marginal $p(t, E)$, evaluated using the optimism, KM divergence, and short-sightedness metrics. We examine t-SNE plots \citep{van2008visualizing} to provide a qualitative performance of the covariates and Kaplan-Meier plots for time and censoring variables in Appendix \ref{app:qualitative}.
    \item \textbf{Downstream Performance:} 
    Section \ref{subsection:experiments_performance} compares the performance of survival models trained with synthetic data generated by \methodname{} to models trained with data generated from the baselines. This is used to quantify the quality of the full $p(\vec x, t, E)$ distribution. Here, a good result is when a model trained with synthetic data performs similarly to one trained with real data or in rare cases better \citep{luo2018eeg}, and outperforms models trained with different synthetic data. 
    \item \textbf{Ablation study:} 
    In Section \ref{subsection:experiments_gains}, we perform an ablation study to demonstrate and quantify the importance of each component of \methodname{}. This provides insight into what each component of \methodname{} offers the overall model, and by using the new time and censoring metrics we are able to determine \emph{how} \methodname{} fails when certain components are missing.
\end{enumerate}

\paragraph{Benchmarks.} We compare \methodname{} against the following benchmarks: generative adversarial networks for anonymization (\textbf{ADS-GAN}) \citep{yoon2020adsgan}; conditional generative adversarial networks for tabular data (\textbf{CTGAN}) \citep{xu2019ctgan}; variational autoencoder for tabular data (\textbf{TVAE}) \citep{xu2019ctgan}; a variant of Bayesian Networks (\textbf{PrivBayes}) \citep{zhang_privbayes}; and Normalizing flows for tabular data (\textbf{NFlows}) \citep{papamakarios_nflows}. For a fair comparison, we preprocess the data using our tabular encoder for all methods that are not specifically adapted to support tabular data (ADS-GAN, Normalizing flows).

\paragraph{Datasets.} We test \methodname{} on a variety of medical survival analysis datasets. The datasets are: (1) ACTG 320 clinical trial dataset (\textbf{AIDS}) \citep{hammer1997controlled}; (2) Cambridge Urology Translational Research and Clinical Trials dataset for prostate cancer mortality in the UK (\textbf{CUTRACT}) \citep{cutract2019}; (3) a private heart failure dataset (\textbf{PHEART}); (4) SEER dataset for prostate cancer mortality in the US (\textbf{SEER}) \citep{seer2019} and (5) The Molecular Taxonomy of Breast Cancer International Consortium dataset (\textbf{METABRIC}) \citep{metabric2016}. Details on each dataset can be found in Appendix \ref{app:datasets}.

\paragraph{Evaluation.} For each dataset, benchmark, and experimental setting, evaluations are performed using 5 different random seeds, and we report the mean and standard deviations of the desired metric. Further experimental details are provided in Appendix \ref{app:experimental_details} and additional experiments in Appendix \ref{app:additional_experiments}.
Code reproducing all experiments and implementing \methodname{} is publicly available at: \url{https://github.com/vanderschaarlab/survivalgan}.

\subsection{Quality of Marginal Distributions}
\label{subsection:experiments_diversity}
\label{subsection:experiments_fidelity}

\paragraph{Covariates.} To evaluate the quality of the covariates, in Table \ref{table:experiments_data_fidelity} we report the values of the Jensen-Shannon distance and Wasserstein distance. We do not expect \methodname{} to produce higher quality covariates than the baselines but we must confirm that they are not significantly worse. We see that \methodname{} provides robust coverage of the covariate space, as well as the benchmarks, often achieving the best or close to the best score.

\begin{table*}[ht!]\scriptsize
\centering
\caption{Predictive performance of discriminative models trained with synthetic data. $*$: The discriminative models failed to train on the generated data. We provide the results when training with the original real data for reference.}
\label{table:experiments_predictive_performance_id}
\begin{widetable}{\textwidth}{c|c|c|c|c|c|c}
\toprule
  Metric & Method & AIDS & CUTRACT & PHEART & SEER & METABRIC \\
\midrule
\multirow{7}{*}{
    \begin{tabular}{c}
        C-Index \\
        \tiny{(Higher is Better)}
    \end{tabular}
}
& SurvivalGAN & 
    \textbf{0.678$\pm$0.03} & 
    \textbf{0.799$\pm$0.02} & 
    \textbf{0.638$\pm$0.01} & 
    \textbf{0.835$\pm$0.01} & 
    \textbf{0.734$\pm$0.01} \\ 
& PrivBayes & 
    0.504$\pm$0.09 & 
    0.544$\pm$0.15 & 
    0.557$\pm$0.01 & 
    0.550$\pm$0.16 & 
    0.334$\pm$0.24 \\ 
& ADS-GAN & 
    0.541$\pm$0.06 & 
    0.607$\pm$0.10 & 
    0.565$\pm$0.03 & 
    $*$ & 
    0.546$\pm$0.03 \\ 
& CTGAN & 
    0.546$\pm$0.08 & 
    0.791$\pm$0.02 & 
    0.612$\pm$0.01 & 
    0.830$\pm$0.01 & 
    0.499$\pm$0.04 \\ 
& TVAE & 
    0.564$\pm$0.04 & 
    0.779$\pm$0.03 & 
    0.621$\pm$0.02 & 
    0.808$\pm$0.02 & 
    0.711$\pm$0.02 \\ 
& NFlows & 
    0.522$\pm$0.08 & 
    0.730$\pm$0.07 & 
    0.566$\pm$0.02 & 
    0.776$\pm$0.04 & 
    0.482$\pm$0.01 \\ 
& Original &
    0.742$\pm$0.02 & 
    0.826$\pm$0.01 & 
    0.668$\pm$0.01 & 
    0.856$\pm$0.01 & 
    0.706$\pm$0.02  
    \\
\midrule
\midrule
\multirow{7}{*}{
    \begin{tabular}{c}
        Brier Score \\
        \tiny{(Lower is Better)}
    \end{tabular}
}
& SurvivalGAN & 
    \textbf{0.057$\pm$0.01} & 
    \textbf{0.084$\pm$0.01} & 
    \textbf{0.181$\pm$0.01} & 
    \textbf{0.023$\pm$0.01} & 
    0.179$\pm$0.01 \\ 
& PrivBayes & 
    0.058$\pm$0.01 & 
    0.100$\pm$0.02 & 
    0.209$\pm$0.02 & 
    0.025$\pm$0.02 & 
    0.562$\pm$0.31 \\ 
& ADS-GAN & 
    0.060$\pm$0.01 & 
    0.117$\pm$0.01 & 
    0.231$\pm$0.03 & 
    $*$ & 
    0.260$\pm$0.05 \\ 
& CTGAN & 
    0.061$\pm$0.01 & 
    0.172$\pm$0.03 & 
    0.188$\pm$0.02 & 
    0.115$\pm$0.03 & 
    0.183$\pm$0.02 \\ 
& TVAE & 
    0.061$\pm$0.02 & 
    0.099$\pm$0.01 & 
    0.206$\pm$0.02 & 
    0.025$\pm$0.01 & 
    \textbf{0.161$\pm$0.01} \\ 
& NFlows & 
    0.097$\pm$0.03 & 
    0.171$\pm$0.04 & 
    0.192$\pm$0.01 & 
    0.164$\pm$0.08 & 
    0.173$\pm$0.01 \\ 
& Original & 
    0.072$\pm$0.01 & 
    0.095$\pm$0.01 & 
    0.166$\pm$0.01 & 
    0.025$\pm$0.01 & 
    0.161$\pm$0.001  
    \\
\bottomrule
\end{widetable}
\end{table*}

\begin{table*}[ht!]\scriptsize
\caption{Source-of-Gain Analysis on Multiple Datasets. $*$: The discriminative model failed to train on the generated data.}
\label{table:experiments_gain_source}
\begin{widetable}{\textwidth}{c|c|c|c|c|c}
\toprule
 Metric & Method & AIDS & CUTRACT & PHEART & SEER \\
\midrule\multirow{5}{*}{
    \begin{tabular}{c}
        C-Index \\
        \tiny{(Higher is Better)}
    \end{tabular}
}
& SurvivalGAN & 
    \textbf{0.723$\pm$0.02} & 
    \textbf{0.804$\pm$0.01} & 
    \textbf{0.644$\pm$0.01} &
    \textbf{0.834$\pm$0.01}  \\
& w/o Time Regressor & 
    0.688$\pm$0.03 & 
    0.719$\pm$0.07 & 
    0.558$\pm$0.02 &
    0.677$\pm$0.02 \\
& w/o Imbalanced Sampling & 
    $*$ & 
    0.671$\pm$0.12 & 
    0.590$\pm$0.02 &
    0.504$\pm$0.01\\
& w/o Temporal Sampling & 
    0.713$\pm$0.04 & 
    0.792$\pm$0.01 & 
    0.614$\pm$0.01 &
    0.636$\pm$0.13 \\
& w/o Cond. GAN & 
    0.714$\pm$0.04 & 
    0.655$\pm$0.13 & 
    0.563$\pm$0.04 &
    0.573$\pm$0.12\\
\midrule
\multirow{5}{*}{
    \begin{tabular}{c}
        Brier Score \\
        \tiny{(Lower is Better)}
    \end{tabular}
}
& SurvivalGAN & 
    \textbf{0.066$\pm$0.01} & 
    \textbf{0.083$\pm$0.01} & 
    \textbf{0.176$\pm$0.01} &  
    \textbf{0.022$\pm$0.01}\\
& w/o Time Regressor & 
    0.144$\pm$0.02 & 
    0.182$\pm$0.02 & 
    0.233$\pm$0.02 &
    0.252$\pm$0.05  \\
& w/o Imbalanced Sampling & 
    $*$ & 
    0.109$\pm$0.01 & 
    0.229$\pm$0.01 &
    0.025$\pm$0.01 \\
& w/o Temporal Sampling & 
    0.152$\pm$0.01 & 
    0.113$\pm$0.01 & 
    0.223$\pm$0.01 &
    0.153$\pm$0.11 \\
& w/o Cond. GAN & 
    0.187$\pm$0.01 & 
    0.247$\pm$0.06 & 
    0.234$\pm$0.04 &
    0.224$\pm$0.11\\
\midrule
\midrule
\multirow{5}{*}{
    \begin{tabular}{c}
        Optimism \\
        \tiny{(Closer to Zero is Better)}
    \end{tabular}
}
& SurvivalGAN &     
    \textbf{-0.006$\pm$0.01} & 
    \textbf{-0.012$\pm$0.03} & 
    \textbf{-0.036$\pm$0.05} & 
    \textbf{-0.079$\pm$0.02} \\
& w/o Time Regressor & 
   0.067$\pm$0.01 &
   -0.025$\pm$0.15 & 
   -0.070$\pm$0.06 & 
   0.123$\pm$0.01\\
& w/o Imbalanced Sampling & 
   0.066$\pm$0.01 & 
   0.051$\pm$0.11 & 
   0.048$\pm$0.16 & 
   0.096$\pm$0.01 \\
& w/o Temporal Sampling & 
   -0.284$\pm$0.07 & 
   -0.266$\pm$0.01 & 
   -0.116$\pm$0.07 & 
   -0.644$\pm$0.03 \\
& w/o Cond. GAN & 
   -0.309$\pm$0.03 & 
   -0.369$\pm$0.01 & 
   	-0.130$\pm$0.04 & 
   -0.445$\pm$0.11 \\
 \midrule
\multirow{5}{*}{
    \begin{tabular}{c}
        Short-sightedness \\
        \tiny{(Closer to Zero is Better)}
    \end{tabular}
}
& SurvivalGAN & 
    \textbf{0.007$\pm$0.01} & 
    \textbf{0.046$\pm$0.05} & 
    \textbf{0.127$\pm$0.15} & 
    \textbf{0.010$\pm$0.04}\\ 
& w/o Time Regressor & 
   0.009$\pm$0.01 &
   0.117$\pm$0.08 & 
   0.497$\pm$0.05 & 
   0.040$\pm$0.01\\
& w/o Imbalanced Sampling & 
   0.008$\pm$0.01 & 
   0.064$\pm$0.06 & 
   0.228$\pm$0.14 & 
   0.132$\pm$0.12 \\
& w/o Temporal Sampling & 
   0.012$\pm$0.01 & 
   0.051$\pm$0.07 & 
   0.128$\pm$0.12 & 
   0.145$\pm$0.05 \\
& w/o Cond. GAN & 
   0.019$\pm$0.02 & 
   0.127$\pm$0.03 & 
   0.085$\pm$0.13 & 
   0.049$\pm$0.01 \\
\bottomrule
\end{widetable}
\end{table*}

\paragraph{Time \& Censoring.} To evaluate $p(t, E)$ we report the optimism, KM divergence, and short-sightedness metrics in Table \ref{table:experiments_data_fidelity}. SurvivalGAN shows stable results across all datasets, consistently achieving the best value or close to the best optimism and KM divergences. On the whole, the majority of baselines do not suffer significantly from short-sightedness. We see that \methodname{} is always close to the best value for a given dataset. Crucially, we see that \methodname{} is never extremely short-sighted. Whereas we see ADS-GAN, CTGAN and TVAE can suffer extreme failure cases measured by short-sightedness. This underlines one of our main novelties: good coverage of both covariate and temporal space by handling the censoring of the data, to robustly generate survival data.

\subsection{Downstream Performance}
\label{subsection:experiments_performance}

To assess downstream performance, we train a set of discriminative models on the synthetic data and test them on the real data, known as the Train on Synthetic Test on Real (TSTR) metric \citep{esteban2017real}.
The discriminative models fall into different categories of survival models: linear models (CoxPH) \citep{cox1972regression}, gradient boosting (SurvivalXGBoost) \citep{barnwal2022survival}, random forests (RandomSurvivalForest) \citep{ishwaran2008random}, and neural networks (DeepHit) \citep{lee2018deephit}.
For each experiment, we report the concordance index (\textbf{C-Index}) \citep{harrell1982evaluating}, a standard metric for assessing the quality of the ranking in survival models, and the \textbf{Brier Score} \citep{brier1950verification}, which measures the calibration of the probabilistic predictions. We evaluate the performance using 3-fold cross-validation on the real data for each discriminative model. The generative models are trained with training data, the predictive models are then trained on the synthetic data and tested on a held-out test set.
Table \ref{table:experiments_predictive_performance_id} shows the downstream performance of the survival models. We select the best-performing predictive model for each metric and report its score. SurvivalGAN consistently leads to better-performing survival models, both in terms of prediction quality (C-Index) and calibration (Brier Score). 

\begin{figure*}[ht!]
\includegraphics[width=\linewidth]{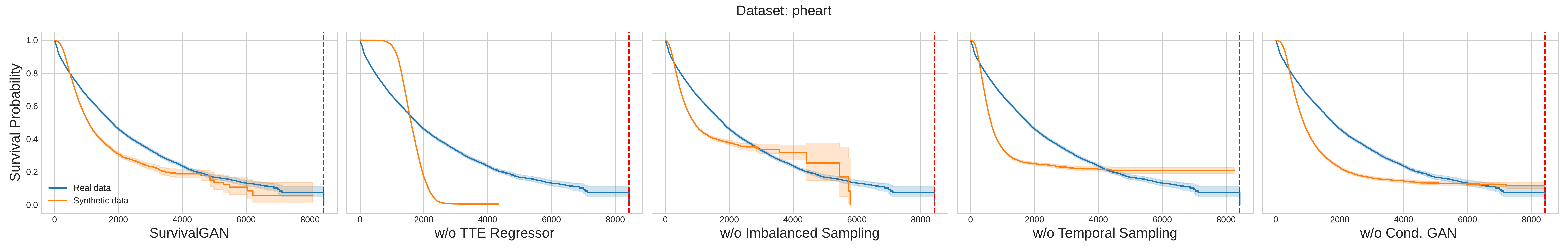}
\caption{Sources of gain visualization for temporal quality using Kaplan-Meier plots. The sightedness is improved by the time regressor ($2^{nd}$ plot) and by the imbalanced sampling ($3^{rd}$ plot). The optimism is improved by the temporal sampling ($4^{th}$ plot) and by the conditional GAN ($5^{th}$ plot).}
\label{fig:experiments_gains_km}
\end{figure*}

\subsection{Sources of Gain}
\label{subsection:experiments_gains}

SurvivalGAN is characterized by: (1) the time-to-event/censoring regressor; (2) the imbalanced sampling with respect to time horizons and censoring and (3) the conditional GAN used for the data generation. To examine the importance of each contribution, we apply the following modifications to SurvivalGAN: (1) with a standard time regressor (DATE) \citep{chapfuwa2018adversarial} instead of ours; (2) without the imbalanced sampling; (3) with imbalanced sampling, but only targeting the censoring instead of censoring and time horizon and (4) without the conditional GAN. We then evaluate downstream performance of predictive models. The ablation study was conducted in an in-distribution manner, leading to minor performance differences with Table \ref{table:experiments_predictive_performance_id}. In addition to predictive performance, we also evaluate optimism and short-sightedness.
We observe in Table \ref{table:experiments_gain_source} that all three components make significant contributions to improving the quality of the generated data.

\paragraph{Insight.} 
The time regressor has an essential role in the quality of the rankings in the survival data (C-Index), as seen in the AIDS and CUTRACT datasets. It is also important with respect to short-sightedness: we see in particular for PHEART that, without a dedicated time regressor, the model is noticeably short-sighted. 
The imbalanced sampler plays a key role in the downstream models' ability to rank samples (measured by C-Index) for all datasets, but notably the large datasets, as seen in PHEART and SEER.
Interestingly, without the temporal sampling or the conditional GAN, the model is severely pessimistic (negative optimism).
Finally, without the conditional GAN, the model typically has a far worse Brier score across all datasets, showing that the conditional GAN significantly improves calibration.

We also visualize these sources of gain qualitatively using KM plots on the PHEART dataset in Figure \ref{fig:experiments_gains_km}. We see that all components of \methodname{} are required for optimal performance. This supports the idea that the imbalanced sampler and time regressor are crucial for the quality of the time values, while the conditional GAN is critical for temporal calibration.

\section{CONCLUSION}
\label{section:conclusion}

We investigated the problem of generating synthetic survival data with censoring. We first formalized the problem, identifying three possible failure modes specific to the survival setting. We then introduced three new metrics based on the survival functions to quantify these phenomena.
On top of this, we introduced \methodname{}, a generative model that generates synthetic survival data. This is accomplished by incorporating censored and non-censored data into the training process, unlocking the use of abundant censored data. Additionally, the time-to-event/censoring data is generated in a more principled way, using a pre-trained survival function and time-to-event model, which permits future extensions to competing events. \methodname{} was tested on multiple medical datasets, generating more faithful data and leading to better downstream models than standard baseline generative methods. 

\paragraph{Limitations.} Currently \methodname{} is not able to address distribution shifts over time, where for example due to advances in medicine we might expect better survival rates in the future than we do now. In addition, it currently only operates in the static setting, with extensions to temporal data future work. Additionally, \methodname{} does not provide \emph{guarantees} on privacy such as those in \citet{jordon2018pate}. We view these as exciting lines of future research.

\section*{Acknowledgements}
We thank the anonymous reviewers for their comments and suggestions. Alexander Norcliffe is supported by a GlaxoSmithKline grant. Fergus Imrie and Mihaela van der Schaar are supported by the National Science Foundation (NSF, grant number 1722516).
Mihaela van der Schaar is additionally supported by the Office of Naval Research (ONR).

\clearpage
\bibliography{main}

\clearpage

\appendix
\onecolumn

\aistatssupplementarytitle{Supplementary Material for: 
\\
SurvivalGAN: Generating time-to-event Data for Survival Analysis}

\section{BROADER IMPACT}
\label{app:broader_impact}

\paragraph{Applications.} In general, there exist malicious applications of generative models. For example, when creating image or speech data, it is possible to make counterfeit images or recordings of an opponent, falsely damaging their reputation. Our paper is on survival data and we do not envision such malicious uses. Survival data is specific to the domain it is being applied to, which limits the possibility of using fake data for unethical purposes. As with all generative models, it is possible to reinforce biases in the training data. However, by providing a condition to the generator we are able to sample from underrepresented groups and our work makes progress in that respect. 

Further, the use of \methodname{} allows us to train survival models without original data, removing real, sensitive human data from the supervised training process. This also makes it possible to generate more data quickly, helping to train survival models, which we have seen have positive societal impacts \citep{gross2014rate, arsene2007artificial, danacica2010using}.

\paragraph{Datasets.} In our experiments we use medical datasets. In general, these can contain sensitive information about participants. We note that we did not collect any data; all data was collected by external labs/medical researchers. All identifiable information has been removed from the datasets by the curators and permission was given by the subjects, making the datasets suitable for this paper.

\section{METRICS}
\label{app:metrics}

\begin{figure}[h]
    \centering
    \includegraphics[width=\textwidth]{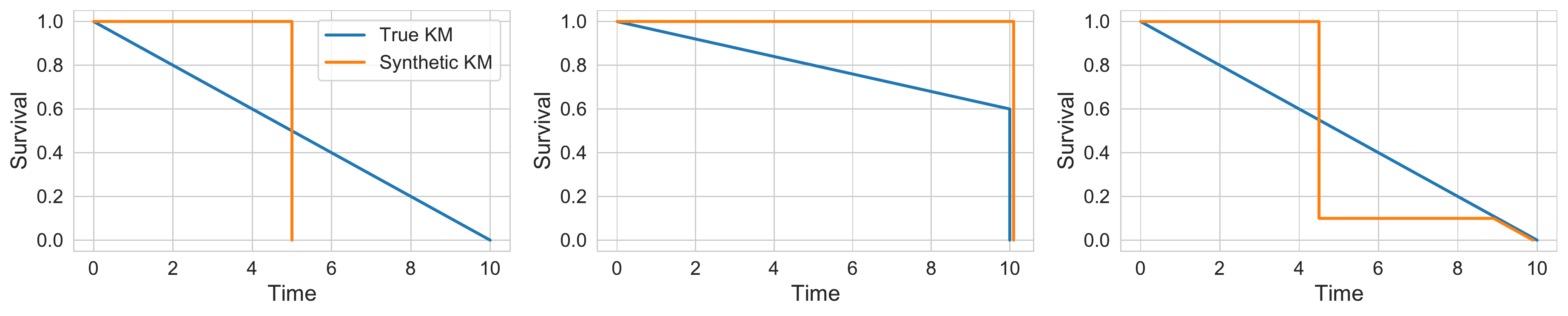}
    \caption{Illustrative examples of Kaplan-Meier plots to show the need for all three metrics. Left: Optimism is zero, but short-sightedness and KM divergence are greater than zero. Middle: Short-sightedness is zero but optimism and KM divergence are greater than zero. Right: Optimism and short-sightedness are both zero, but KM divergence is greater than zero.}
    \label{fig:examples}
\end{figure}

In Section \ref{sec:metrics}, we introduced three new metrics: optimism, short-sightedness, and Kaplan-Meier divergence. These metrics capture different nuances in generating data for survival analysis. In particular, we can see that we require three metrics because it is possible for one value to be zero, while the other two are non-zero. Figure \ref{fig:examples} provides an illustration of this, showing the Kaplan-Meier plots for different situations and how the metrics differ. While KM divergence is enough to indicate when synthetic data is inadequate, it does not tell us \emph{how}, for this we need the other two metrics because they target specific failure modes when modeling $p(t, E)$, giving us an interpretable meaning. Note that a KM divergence of zero means that optimism and short-sightedness must also both be zero.

\paragraph{Bounds on Optimism.} Optimism is trivially bounded by -1 and 1 from its definition. Here we demonstrate that it is possible to bound the optimism by twice the total variation between the synthetic and real distributions $p_{\text{Syn}}(t)$ and $p_{\text{Real}}(t)$; which may be a tighter bound than -1 and 1 in certain situations. Recall that $p(t)$ is the probability density that the event of interest happens at time $t$, and the survival function is defined as $S(t) = \int_{t}^{\infty}p(t')dt'$. These can both be conditioned on covariate $\vec x$ which is omitted here for clarity. Differentiating this expression with respect to $t$ we obtain

\[
\frac{dS}{dt} = -p(t).
\]

%

Now we look at the definition of optimism

\[
\text{Optimism} = \frac{1}{T}\int_{0}^{T}
\big(
S_{\text{Syn}}(t) - S_{\text{Real}}(t)
\big) 
dt.
\]

Applying integration by parts we get
\[
\text{Optimism} = 
\bigg(
S_{\text{Syn}}(T) - S_{\text{Real}}(T)
\bigg)
+ \int_{0}^{T}\frac{t}{T}
\big(
p_{\text{Syn}}(t) - p_{\text{Real}}(t)
\big)
dt.
\]

Rewriting the first bracketed term using the definition of the survival function we get
\[
\text{Optimism} = \int_{T}^{\infty}1\times 
\big(
p_{\text{Syn}}(t) - p_{\text{Real}}(t)
\big)
dt
+
\int_{0}^{T}\frac{t}{T} \times
\big(
p_{\text{Syn}}(t) - p_{\text{Real}}(t)
\big)
dt.
\]

We then define $f(t)$ piecewise as
\[
f(t) = 
\begin{cases}
t/T, \qquad  \text{for} ~~~ 0\leq t \leq T \\
1, \qquad \quad  \text{for} ~~~ t > T \\
\end{cases}.
\]

Giving the optimism as a single integral
\[
\text{Optimism} = \int_{0}^{\infty}f(t)
\big(
p_{\text{Syn}}(t) - p_{\text{Real}}(t)
\big)
dt.
\]

We then use the fact that $f(t) \leq 1$ and $(p_{\text{Syn}}(t) - p_{\text{Real}}(t)) \leq |p_{\text{Syn}}(t) - p_{\text{Real}}(t)|$ to conclude that
\[
\text{Optimism} < 2\int_{0}^{\infty}\frac{1}{2}|p_{\text{Syn}}(t) - p_{\text{Real}}(t)| dt.
\]

The strict inequality comes from the fact that $f(t) < 1$ for $t<T$. The integral is the definition of the total-variation divergence. We obtain the lower bound by considering the negative of the above step, that is $(p_{\text{Syn}}(t) - p_{\text{Real}}(t)) \geq -|p_{\text{Syn}}(t) - p_{\text{Real}}(t)|$. Giving us

\begin{equation}
    -2\mathcal{D}_{TV}(p_{\text{Real}}||p_{\text{Syn}}) 
    < \text{Optimism} < 2\mathcal{D}_{TV}(p_{\text{Real}}||p_{\text{Syn}}).
\end{equation}

This allows us to apply other known bounds between different probability distances and the total variation. For example, we can use Pinkser's inequality \citep{csiszar2011information} to bound Optimism with the Kullback-Leibler divergence

\begin{equation}
\label{eq:kl_bound1}
-\sqrt{2\mathcal{D}_{KL}(p_{\text{Real}}||p_{\text{Syn}})}
< \text{Optimism} <  \sqrt{2\mathcal{D}_{KL}(p_{\text{Real}}||p_{\text{Syn}})}.
\end{equation}

We can also apply the inequality of Bretagnolle and Huber \citep{bretagnolle1978estimation, tsybakov2009introduction}

\begin{equation}
\label{eq:kl_bound2}
-2\sqrt{1 - \exp(-\mathcal{D}_{KL}(p_{\text{Real}}||p_{\text{Syn}}))}
< \text{Optimism} <  
2\sqrt{1 - \exp(-\mathcal{D}_{KL}(p_{\text{Real}}||p_{\text{Syn}}))
}.
\end{equation}

Another example is with the Hellinger distance \citep{harsha2007communication}

\begin{equation}
\label{eq:hellinger}
-2\sqrt{2}\mathcal{D}_{H}(p_{\text{Real}}||p_{\text{Syn}})
< \text{Optimism} < 2\sqrt{2}\mathcal{D}_{H}(p_{\text{Real}}||p_{\text{Syn}}).
\end{equation}

The total-variation distance is symmetric. Therefore, despite the Kullback-Leibler divergence and Hellinger distance not being symmetric themselves, we are able to swap $p_{\text{Real}}$ and $p_{\text{Syn}}$ in Equations (\ref{eq:kl_bound1}), (\ref{eq:kl_bound2}) and (\ref{eq:hellinger}) to obtain the tightest bounds. In specific situations there may be a closed-form solution for some of these divergences, allowing us to quickly establish bounds on the optimism.

\section{EXPERIMENTAL DETAILS}
\label{app:experimental_details}

In this section we give full experimental details, giving the dataset descriptions and hyperparameters.

\paragraph{Reproducibility.} All hyperparameters are given in this section for reproducibility. Three of our datasets (SEER, AIDS and METABRIC) are public, making those experiments straightforward to run. It is possible to run all experiments with limited compute. Ours were run locally on a machine with 32GB RAM, Intel Core i7-6700 HQ, GeForce GTX 950M.

\subsection{Data Description}
\label{app:datasets}

In Table \ref{table:survivalgan_evaluation_datasets} we provide details about the datasets used in our evaluation. Of the datasets, AIDS, SEER and METABRIC are public, CUTRACT and PHEART are licensed. AIDS contains people with HIV and SEER with Prostate Cancer. CUTRACT is owned by Cambridge Prostate Cancer (\url{https://cambridgeprostatecancer.com/})\footnote{ Terms and Conditions: \\ \url{https://cambridgeprostatecancer.com/terms-privacy-policy-gdpr-cookies/}}, and focuses on patients with Prostate Cancer. PHEART consists  of patients from 30 medical studies who have experienced heart failure. The Molecular Taxonomy of Breast Cancer International Consortium (METABRIC) database is a Canada-UK Project which contains targeted sequencing data of primary breast cancer samples, this version contains samples 1,980 and 689 features \citep{metabric2016}.

\begin{table}[ht!]\scriptsize
\centering
\caption{Datasets used for evaluation.}
\label{table:survivalgan_evaluation_datasets}
\begin{tabular}{ccccc}
\toprule
\textbf{Dataset} & \textbf{No. instances} & \textbf{No. censored instances} & \textbf{No. features} & \textbf{Experiment label} \\ 
\midrule
ACTG 320 clinical trial dataset & 1151 & 1055 & 11 & AIDS \\
\midrule
CUTRACT & 10086 & 8881 & 6 & CUTRACT \\ 
\midrule
PHEART & 40409 & 25664 & 29 & PHEART \\ 
\midrule
SEER prostate cancer& 171942 & 167568 & 6 & SEER \\
\midrule
METABRIC & 1093 & 609 & 689 & METABRIC
\\
\bottomrule
\end{tabular}
\end{table}

\paragraph{Evaluation Horizons.}
The evaluation process is averaged over 5 time horizons, specific to each dataset. Given time-to-event $T$ in the training set, we select the 5 evaluation horizons, evenly spaced between $[T.min(), T.max()]$. Table \ref{tab:evaluation_horizons} gives the evaluation horizons for each dataset.

\begin{table}[ht!]\scriptsize
\centering
\caption{Time horizons used for evaluation by dataset (can represent days/months, depending on the datasets).}
\begin{tabular}{cccccc}
\toprule
\textbf{Dataset} & \textbf{Horizon 1} & \textbf{Horizon 2} & \textbf{Horizon 3} & \textbf{Horizon 4} & \textbf{Horizon 5} \\ \midrule
AIDS (Months) & 61.5 & 122. & 182.5 & 243. & 303.5 \\ 
\midrule
CUTRACT  (Days) & 1051. & 2082. & 3113. & 4144. & 5175. \\ 
\midrule
PHEART (Days) & 1404.6 & 2809.3 & 4214. & 5618.6 & 7023.3 \\ 
\midrule
SEER (Days) & 775. & 1550. & 2325. & 3100. & 3875. \\ 
\midrule
METABRIC (Months) & 56.25 & 112.41 & 168.56 & 224.72 & 280.87 \\
\bottomrule
\end{tabular}
\label{tab:evaluation_horizons}
\end{table}

\clearpage
\subsection{Hyperparameters}

\label{app:hyperparams}

In Table \ref{tab:survivalgan_hyperparams_domain}, we present the full configuration of \methodname{} used in our experiments. Table \ref{tab:tte_hyperparams_domain} contains the hyperparameters used by the predictive models trained with synthetic data for downstream tasks. Finally, Table \ref{tab:synth_benchmarks_hyperparams_domain} details the hyperparameters used for the synthetic benchmarks.

\begin{table}[ht!]
\centering
\caption{SurvivalGAN Hyperparameters by Component.}
\label{tab:survivalgan_hyperparams_domain}
\begin{widetable}{\textwidth}{lll}
\toprule
 \textbf{Component} & \textbf{Parameter} & \textbf{Parameter Value} \\
\midrule\multirow{11}{*}{Survival Function} 
    & Model & Deephit \\
    & No. Durations & $100$ \\
    & Batch Size & $100$ \\
    & No. Epochs & $2000$ \\
    & Learning Rate & $1\times10^{-3}$ \\
    & Hidden Width & $300$ \\
    & $\alpha$ & $0.28$ \\
    & $\sigma$ & $0.38$ \\
    & Dropout Rate & $0.02$ \\
    & Patience & $20$ \\
    & Using Batch Normalization & True \\
\midrule\multirow{5}{*}{Time-to-event Regressor} 
    & Model & XGBoostRegressor \\
    & No. Estimators & $200$ \\
    & Depth & $5$ \\
    & Booster & Dart \\
    & Tree Method & Histogram \\
\midrule\multirow{15}{*}{Conditional GAN} 
    & Data Sampling Strategy & With Data Frequencies
    \\
    & No. Iterations & $1500$ \\
    & Generator No. Hidden Layers & $3$ \\
    & Generator Hidden Width & $250$ \\
    & Generator Non-linearity & Tanh \\
    & Generator Dropout Rate & $0.1$ \\
    & Discriminator No. Hidden Layers & $2$ \\
    & Discriminator Hidden Width & $250$ \\
    & Discriminator Non-linearity & Leaky ReLU \\
    & Discriminator Dropout Rate & $0.1$ \\
    & Learning Rate & $1\times10^{-3}$ \\
    & Weight Decay & $1\times10^{-3}$ \\
    & Batch Size & $500$ \\
    & Gradient Penalty ($\lambda$) & $10$ \\
    & Encoder Max Clusters & $10$ \\
\bottomrule
\end{widetable}
\end{table}

\begin{table}[ht!]
\centering
\caption{Hyperparameters used for the baseline time-to-event benchmarks, used as downstream models.}
\label{tab:tte_hyperparams_domain}
\begin{widetable}{\textwidth}{lll}
\toprule
\textbf{Method} & \textbf{Parameter} & \textbf{Parameter Value} \\
\midrule\multirow{3}{*}{CoxPH} 
    & Estimation Method & Breslow \\
    & Penalizer & $0.0$ \\
    & $L^1$ Ratio & $0.0$ \\
\midrule\multirow{3}{*}{Weibull AFT} 
    & $\alpha$ & $0.05$ \\
    & Penalizer & $0.0$ \\
    & $L^1$ Ratio & $0.0$ \\
\midrule\multirow{12}{*}{SurvivalXGBoost} 
    & Objective & Survival: AFT \\
    & Evaluation Metric & AFT Negative Log Likelihood \\
    & AFT Loss Distribution & Normal \\
    & AFT Loss Distribution Scale & $1.0$ \\
    & No. Estimators & $100$ \\
    & Column Subsample Ratio (by node) & $0.5$ \\
    & Maximum Depth & $8$ \\
    & Subsample Ratio & $0.5$ \\
    & Learning Rate & $5\times10^{-2}$ \\
    & Minimum Child Weight & $50$ \\
    & Tree Method & Histogram \\
    & Booster & Dart \\
\midrule\multirow{3}{*}{RandomSurvivalForest} 
    & Max Depth & $3$ \\
    & No. Estimators & $100$ \\
    & Criterion & Gini \\
\midrule\multirow{10}{*}{Deephit} 
    & No. Durations: & $1000$ \\
    & Batch Size & $100$ \\
    & Epochs & $2000$ \\
    & Learning Rate & $1\times10^{-3}$ \\
    & Hidden Width & $300$ \\
    & $\alpha$ & $0.28$ \\
    & $\sigma$ & $0.38$ \\
    & Dropout Rate & $0.02$ \\
    & Patience & $20$ \\
    & Using Batch Normalization & True \\
\midrule\multirow{19}{*}{DATE} 
    & Generator No. Hidden Layers & $2$ \\
    & Generator Hidden Width & $250$ \\
    & Generator Non-linearity & Leaky ReLU \\
    & Generator No. Iterations & $1000$ \\
    & Generator Using Batch Normalization & False \\
    & Generator Dropout Rate & $0.0$ \\
    & Generator Learning Rate & $2\times10^{-4}$ \\
    & Generator Weight Decay & $1\times10^{-3}$ \\
    & Generator Residual & True \\
    & Discriminator No. Hidden Layers & $3$ \\
    & Discriminator Hidden Width & $300$ \\
    & Discriminator Non-linearity & Leaky ReLU \\
    & Discriminator No. Iterations & $1$ \\
    & Discriminator Using Batch Normalization & False \\
    & Discriminator Dropout Rate & 0.1 \\
    & Discriminator Learning Rate & $2\times10^{-4}$ \\
    & Discriminator Weight Decay & $1\times10^{-3}$ \\
    & Patience & $10$ \\
    & Batch Size & $100$ \\
\bottomrule
\end{widetable}
\end{table}

\begin{table}[h!]
\centering
\caption{Hyperparameters for the synthetic benchmarks.}
\label{tab:synth_benchmarks_hyperparams_domain}
\begin{widetable}{\textwidth}{lll}
\toprule
\textbf{Method} & \textbf{Parameter} & \textbf{Parameter Value} \\
\midrule\multirow{3}{*}{PrivBayes} 
    & $\epsilon$ & $1.0$  \\
    & $\theta$ Usefulness & $4$  \\
    & $\epsilon$ Split & $0.3$  \\
\midrule\multirow{15}{*}{ADS-GAN} 
    & No. Iterations & $1500$ \\
    & Generator no. Hidden Layers & $3$ \\
    & Generator Hidden Width & $250$ \\
    & Generator Non-linearity & Tanh \\
    & Generator Dropout Rate & $0.1$ \\
    & Discriminator No. Hidden Layers & $2$ \\
    & Discriminator Hidden Width & $250$ \\
    & Discriminator Non-linearity & Leaky ReLU \\
    & Discriminator Dropout Rate & $0.1$ \\
    & Learning Rate & $1\times10^{-3}$ \\
    & Weight Decay & $1\times10^{-3}$ \\
    & Batch Size & $500$ \\
    & Gradient Penalty ($\lambda$) & $10$ \\
    & Identifiability Penalty & $0.1$ \\
    & Encoder Max Clusters & $10$ \\
\midrule\multirow{13}{*}{CTGAN} 
    & Embedding Width & $10$  \\
    & Generator No. Hidden Layers & $2$  \\
    & Generator Hidden Width & $256$  \\
    & Generator Learning Rate & $2\times10^{-4}$  \\
    & Generator Decay & $1\times10^{-6}$  \\
    & Discriminator No. Hidden Layers & $2$  \\
    & Discriminator Hidden Width & $256$  \\
    & Discriminator Learning Rate & $2\times10^{-4}$  \\
    & Discriminator Decay & $1\times10^{-6}$  \\
    & Batch Size & $500$  \\
    & Discriminator Steps & $1$  \\
    & No. Iterations & $300$  \\
    & Pac & $10$  \\
\midrule\multirow{9}{*}{TVAE} 
    & Embedding Width & $128$ \\
    & Encoder No. Hidden Layers & $2$ \\
    & Encoder Hidden Width & $128$ \\
    & Decoder No. Hidden Layers & $2$ \\
    & Decoder Hidden Width & $128$ \\
    & $L^2$ Scale & $1\times10^{-5}$ \\
    & Batch Size & $500$ \\
    & No. Iterations & $300$ \\
    & Loss Factor & $2$ \\
\midrule\multirow{12}{*}{NFlows} 
    & No. Iterations & $500$  \\
    & No. Hidden Layers & $1$  \\
    & Hidden Width & $100$  \\
    & Batch Size & $100$  \\
    & No. Transform Blocks & $1$  \\
    & Dropout Rate & $0.1$  \\
    & No. Bins & $8$  \\
    & Tail Bound & $3$  \\
    & Learning Rate & $1\times10^{-3}$  \\
    & Base Distribution & Standard Normal  \\
    & Linear Transform Type & Permutation  \\
    & Base Transform Type & Affine-Coupling  \\
\bottomrule
\end{widetable}
\end{table}

\clearpage
\newpage

\section{ADDITIONAL EXPERIMENTS}
\label{app:additional_experiments}

Here we present extended results from the experiments in the main paper. We first conduct further evaluation of the covariates. We then look at three further ablations of \methodname{}:  using a dedicated censoring network; replacing the conditional GAN with a conditional VAE and replacing the GAN with a Gaussian mixture model. Finally, we provide qualitative results in the form of t-SNE and Kaplan-Meier plots in Appendix \ref{app:qualitative}.

\subsection{Statistical Metrics for the Covariates}

In Table \ref{table:experiments_data_fidelity}, we reported the Jensen-Shannon distance and Wasserstein distance of synthetic covariates compared to real covariates. We saw that \methodname{} generates data close to the real distribution, either similar to or better than the baselines. Here we extend this and record four further metrics:
\begin{enumerate}[leftmargin=*, noitemsep,topsep=0pt]
    \item The Precision \citep{sajjadi2018assessing} - measures the rate by which the generative model synthesizes realistic-looking samples, higher is better.
    \item The Recall \citep{sajjadi2018assessing} - measures the fraction of real samples that are covered by the synthetic data, higher is better. 
    \item The Density \citep{naeem2020reliable} - improves the precision metric and measures how many real-sample neighborhood spheres contain the generated data, higher is better.
    \item The Coverage \citep{naeem2020reliable} - improves the recall metric and reports the fraction of real samples whose neighborhoods contain at least one generated sample, higher is better.
\end{enumerate}

We give these values in Table \ref{table:further_covariates}. We see that \methodname{} again performs on par with the baselines. This shows that \methodname{} does not generate worse covariates than the baselines, which is not the main aim of this work, but is crucial to generate a full synthetic survival dataset.
\clearpage

\begin{table}[hbt!]
\caption{Precision, recall, density, and coverage given the covariates manifolds, higher values are better. The SEER dataset evaluation failed due to memory limits. $*$: Not evaluated.}
\label{table:further_covariates}
\begin{widetable}{\textwidth}{c|c|c|c|c|c}
\toprule
\textbf{Metric} & \textbf{Method} & \textbf{AIDS}  & \textbf{CUTRACT} & \textbf{PHEART} & \textbf{METABRIC}  \\
\midrule
\multirow{6}*{
    \begin{tabular}{c}
        Precision \\
        \tiny{(Higher is Better)}
    \end{tabular}
}
& SurvivalGAN & 
    \multicolumn{1}{r|}{0.964$\pm$0.01} & 
    \multicolumn{1}{r|}{0.986$\pm$0.01} & 
    \multicolumn{1}{r|}{0.958$\pm$0.02} &
    \multicolumn{1}{r}{0.009$\pm$0.01} \\
& PrivBayes & 
    \multicolumn{1}{r|}{0.885$\pm$0.02} & 
    \multicolumn{1}{r|}{0.962$\pm$0.01} & 
    \multicolumn{1}{r|}{0.811$\pm$0.01} &
    \multicolumn{1}{c}{$*$}  \\
& ADS-GAN & 
    \multicolumn{1}{r|}{0.898$\pm$0.13} & 
    \multicolumn{1}{r|}{\textbf{0.989$\pm$0.01}} & 
    \multicolumn{1}{r|}{0.941$\pm$0.04} &
    \multicolumn{1}{r}{0.001$\pm$0.01} \\
& CTGAN & 
    \multicolumn{1}{r|}{0.878$\pm$0.04} & 
    \multicolumn{1}{r|}{0.976$\pm$0.01} & 
    \multicolumn{1}{r|}{0.929$\pm$0.03} &
    \multicolumn{1}{r}{0.006$\pm$0.01}  \\
& TVAE & 
    \multicolumn{1}{r|}{\textbf{0.970$\pm$0.01}} & 
    \multicolumn{1}{r|}{0.960$\pm$0.02} & 
    \multicolumn{1}{r|}{\textbf{0.987$\pm$0.01}} &
    \multicolumn{1}{r}{\textbf{0.014$\pm$0.01}} \\
& NFlows & 
    \multicolumn{1}{r|}{0.880$\pm$0.03} & 
    \multicolumn{1}{r|}{0.864$\pm$0.02} & 
    \multicolumn{1}{r|}{0.582$\pm$0.02} &
    \multicolumn{1}{r}{0.004$\pm$0.01}  \\
    
\midrule
\multirow{6}*{
    \begin{tabular}{c}
        Recall \\
        \tiny{(Higher is Better)}
    \end{tabular}
}
& SurvivalGAN & 
    \multicolumn{1}{r|}{0.911$\pm$0.02} & 
    \multicolumn{1}{r|}{0.714$\pm$0.10} & 
    \multicolumn{1}{r|}{0.689$\pm$0.06} &
    \multicolumn{1}{r}{0.698$\pm$0.04}  \\
& PrivBayes & 
    \multicolumn{1}{r|}{0.968$\pm$0.01} & 
    \multicolumn{1}{r|}{\textbf{0.981$\pm$0.01}} & 
    \multicolumn{1}{r|}{0.975$\pm$0.01} &
    \multicolumn{1}{c}{$*$}  \\
& ADS-GAN & 
    \multicolumn{1}{r|}{0.829$\pm$0.10} & 
    \multicolumn{1}{r|}{0.621$\pm$0.10} & 
    \multicolumn{1}{r|}{0.489$\pm$0.08} &
    \multicolumn{1}{r}{\textbf{0.982$\pm$0.02}} \\
& CTGAN & 
    \multicolumn{1}{r|}{0.947$\pm$0.01} & 
    \multicolumn{1}{r|}{0.980$\pm$0.01} & 
    \multicolumn{1}{r|}{0.931$\pm$0.02} &
    \multicolumn{1}{r}{0.914$\pm$0.06}  \\
& TVAE & 
    \multicolumn{1}{r|}{0.712$\pm$0.07} & 
    \multicolumn{1}{r|}{0.798$\pm$0.02} & 
    \multicolumn{1}{r|}{0.665$\pm$0.01} &
    \multicolumn{1}{r}{0.762$\pm$0.14} \\
& NFlows & 
    \multicolumn{1}{r|}{\textbf{0.972$\pm$0.02}} & 
    \multicolumn{1}{r|}{0.967$\pm$0.01} & 
    \multicolumn{1}{r|}{\textbf{0.991$\pm$0.01}} &
    \multicolumn{1}{r}{0.950$\pm$0.03}  \\
    
\midrule
\multirow{6}*{
    \begin{tabular}{c}
        Density \\
        \tiny{(Higher is Better)}
    \end{tabular}
}
& SurvivalGAN & 
    \multicolumn{1}{r|}{1.018$\pm$0.06} & 
    \multicolumn{1}{r|}{0.976$\pm$0.04} & 
    \multicolumn{1}{r|}{0.902$\pm$0.05} &
    \multicolumn{1}{r}{0.009$\pm$0.01}  \\
& PrivBayes & 
    \multicolumn{1}{r|}{0.719$\pm$0.03} & 
    \multicolumn{1}{r|}{0.876$\pm$0.01} & 
    \multicolumn{1}{r|}{0.557$\pm$0.02} &
    \multicolumn{1}{c}{$*$}  \\
& ADS-GAN & 
    \multicolumn{1}{r|}{0.991$\pm$0.25} & 
    \multicolumn{1}{r|}{1.001$\pm$0.10} & 
    \multicolumn{1}{r|}{ 0.889$\pm$0.10} &
    \multicolumn{1}{r}{0.001$\pm$0.01} \\
& CTGAN & 
    \multicolumn{1}{r|}{0.758$\pm$0.07} & 
    \multicolumn{1}{r|}{0.940$\pm$0.02} & 
    \multicolumn{1}{r|}{0.823$\pm$0.08} &
    \multicolumn{1}{r}{0.003$\pm$0.01}  \\
& TVAE & 
    \multicolumn{1}{r|}{\textbf{1.179$\pm$0.03}} & 
    \multicolumn{1}{r|}{\textbf{1.107$\pm$0.01}} & 
    \multicolumn{1}{r|}{\textbf{1.264$\pm$0.02}} &
    \multicolumn{1}{r}{\textbf{0.013$\pm$0.01}} \\
& NFlows & 
    \multicolumn{1}{r|}{0.641$\pm$0.09} & 
    \multicolumn{1}{r|}{0.605$\pm$0.03} & 
    \multicolumn{1}{r|}{0.292$\pm$0.01} &
    \multicolumn{1}{r}{0.001$\pm$0.01}  \\
    
\midrule
\multirow{6}*{
    \begin{tabular}{c}
        Coverage \\
        \tiny{(Higher is Better)}
    \end{tabular}
}
& SurvivalGAN & 
    \multicolumn{1}{r|}{\textbf{0.919$\pm$0.02}} & 
    \multicolumn{1}{r|}{0.505$\pm$0.06} & 
    \multicolumn{1}{r|}{0.546$\pm$0.01} &
    \multicolumn{1}{r}{0.034$\pm$0.02}  \\
& PrivBayes & 
    \multicolumn{1}{r|}{0.834$\pm$0.04} & 
    \multicolumn{1}{r|}{0.903$\pm$0.01} & 
    \multicolumn{1}{r|}{0.685$\pm$0.01} &
    \multicolumn{1}{c}{$*$}  \\
& ADS-GAN & 
    \multicolumn{1}{r|}{0.792$\pm$0.09} & 
    \multicolumn{1}{r|}{0.472$\pm$0.02} & 
    \multicolumn{1}{r|}{0.493$\pm$0.05} &
    \multicolumn{1}{r}{0.002$\pm$0.01} \\
& CTGAN & 
    \multicolumn{1}{r|}{0.845$\pm$0.05} & 
    \multicolumn{1}{r|}{\textbf{0.927$\pm$0.01}} & 
    \multicolumn{1}{r|}{\textbf{0.874$\pm$0.02}} &
    \multicolumn{1}{r}{0.011$\pm$0.01}  \\
& TVAE & 
    \multicolumn{1}{r|}{0.825$\pm$0.04} & 
    \multicolumn{1}{r|}{0.742$\pm$0.03} & 
    \multicolumn{1}{r|}{0.638$\pm$0.03} &
    \multicolumn{1}{r}{\textbf{0.049$\pm$0.03}} \\
& NFlows & 
    \multicolumn{1}{r|}{0.669$\pm$0.07} & 
    \multicolumn{1}{r|}{0.582$\pm$0.05} & 
    \multicolumn{1}{r|}{0.261$\pm$0.04} &
    \multicolumn{1}{r}{0.006$\pm$0.01}  \\
\bottomrule
\end{widetable}
\end{table}

We additionally evaluate the quality of the covariates by looking at the negative log-likelihood of the synthetic covariates compared to that of the true covariates in Table \ref{tab:metrics_neg_loglikelihood}. We see that \methodname{} typically matches the real data well, often the best or close to best, and never fails significantly, whereas the baselines occasionally contain noticeable failure cases, in particular CTGAN, TVAE and NFlows.

\begin{table}[h]
\caption{Negative log-likelihood in the presence of the covariates. The closer to the real data the better. The values \textbf{closest} to the real data are given in bold. Extreme \underline{failure} cases are underlined. $*$: Not evaluated.}
\centering
\resizebox{\textwidth}{!}{
\begin{tabular}{cccccc}
\toprule
\multicolumn{1}{l|}{\textbf{Source}} & \multicolumn{1}{c|}{\textbf{AIDS} $(/10^{2})$} & \multicolumn{1}{c|}{\textbf{CUTRACT} $(/10^{3})$} & \multicolumn{1}{c|}{\textbf{PHEART} $(/10^{5})$} & \multicolumn{1}{c|}{\textbf{SEER} $(/10^{4})$} & \multicolumn{1}{c}{\textbf{METABRIC} $(/10^{3})$} \\
\midrule
\multicolumn{1}{l|}{Real data} & \multicolumn{1}{c|}{6.21} & \multicolumn{1}{c|}{9.79} & \multicolumn{1}{c|}{1.40} & \multicolumn{1}{c|}{4.66} & \multicolumn{1}{c}{2.02} \\
\midrule
\multicolumn{1}{l|}{SurvivalGAN} & \multicolumn{1}{c|}{\textbf{6.37$\pm$0.72}} & \multicolumn{1}{c|}{11.24$\pm$4.51} & \multicolumn{1}{c|}{\textbf{1.40$\pm$0.25}} & \multicolumn{1}{c|}{\textbf{4.41$\pm$2.21}} & \multicolumn{1}{c}{\textbf{1.35$\pm$0.14}} \\
\multicolumn{1}{l|}{PrivBayes} & \multicolumn{1}{c|}{4.53$\pm$2.83} & \multicolumn{1}{c|}{\textbf{10.31$\pm$0.25}} & \multicolumn{1}{c|}{\textbf{1.41$\pm$0.18}} & \multicolumn{1}{c|}{\textbf{4.84$\pm$0.08}} & \multicolumn{1}{c}{$*$} \\ 
\multicolumn{1}{l|}{CTGAN} & \multicolumn{1}{c|}{10.04$\pm$1.04} & \multicolumn{1}{c|}{\underline{47.72$\pm$1.38}} & \multicolumn{1}{c|}{2.03$\pm$0.02} & \multicolumn{1}{c|}{\underline{99.95$\pm$16.05}} & \multicolumn{1}{c}{1.19$\pm$0.19} \\ 
\multicolumn{1}{l|}{TVAE} & \multicolumn{1}{c|}{\underline{0.12$\pm$0.14}} & \multicolumn{1}{c|}{\underline{2.98$\pm$0.73}} & \multicolumn{1}{c|}{1.31$\pm$0.13} & \multicolumn{1}{c|}{\underline{0.27$\pm$0.15}} & \multicolumn{1}{c}{1.20$\pm$0.37} \\ 
\multicolumn{1}{l|}{NFlows} & \multicolumn{1}{c|}{\underline{26.42$\pm$3.78}} & \multicolumn{1}{c|}{\underline{55.33$\pm$1.82}} & \multicolumn{1}{c|}{1.97$\pm$0.39} & \multicolumn{1}{c|}{\underline{93.79$\pm$30.93}} & \multicolumn{1}{c}{$*$} \\ 
\bottomrule
\end{tabular}}
\label{tab:metrics_neg_loglikelihood}
\end{table}

\clearpage
\subsection{Censoring network}

As a further ablation, we aim to investigate if a dedicated censoring network would improve the quality of SurvivalGAN.
We design the following experiment: We use an XGBoost classifier - denoted Censoring network - to predict the censored/not censored status, based on the covariates. We keep our mechanisms in place: unbalanced time/censoring sampling, and time-to-event/censoring regression. The results are given in Tables \ref{tab:censoring_network_baseline}, \ref{tab:censoring_network_ratio} and \ref{tab:censoring_network_performance}. We see that on the whole, \methodname{} performs better without the censoring network.

\begin{table}[ht!]
\centering
\caption{Censoring network training performance. We evaluate the classifier using only the real data and we report the AUROC. We observe that the classifier has a good performance for distinguishing the classes, on the evaluation datasets.}
\begin{tabular}{cc}
\toprule
\textbf{Dataset} & \textbf{AUROC}\\
\midrule
AIDS & 0.719$\pm$0.037\\
CUTRACT & 0.753$\pm$0.004\\
PHEART & 0.719$\pm$0.004\\ 
SEER & 0.837$\pm$0.003\\ 
METABRIC & 0.717$\pm$0.015 \\
\bottomrule
\end{tabular}
\label{tab:censoring_network_baseline}
\end{table}

\begin{table}[ht!]
\centering
\caption{Number of censored/not censored rows, from the real data, SurvivalGAN, and the censoring network. We want the numbers from the generative models to be as close as possible to those from the real data. While the predictive performance is good overall, there are scenarios like for the AIDS or CUTRACT datasets, where the number of non-censored synthetic subjects from the Censoring Network is too low.}
\begin{tabular}{c|cc|cc|cc}
\toprule
\multirow{2}{*}{\textbf{Dataset}} & \multicolumn{2}{c|}{\textbf{Real data}} & \multicolumn{2}{c|}{\textbf{SurvivalGAN}} & \multicolumn{2}{c}{\textbf{Censoring network}} \\ 
& \multicolumn{1}{c}{\textbf{Censored}} & \textbf{Event} & \multicolumn{1}{c}{\textbf{Censored}} & \textbf{Event} & \multicolumn{1}{c}{\textbf{Censored}} & \textbf{Event} \\
\midrule
AIDS & \multicolumn{1}{c}{1055} & 96 & \multicolumn{1}{c}{1045} & 106 & \multicolumn{1}{c}{1134} & 17 \\ 
CUTRACT & \multicolumn{1}{c}{8881} & 1205 & \multicolumn{1}{c}{8868} & 1218 & \multicolumn{1}{c}{9926} & 160 \\
MAGGIC & \multicolumn{1}{c}{25664} & 14745 & \multicolumn{1}{c}{26019} & 14390 & \multicolumn{1}{c}{27282} & 13127 \\
SEER & \multicolumn{1}{c}{167568} & 4374 & \multicolumn{1}{c}{168866} & 3076 & \multicolumn{1}{c}{132260} & 4872 \\ 
METABRIC & \multicolumn{1}{c}{609} & 484 & \multicolumn{1}{c}{817} & 276 & \multicolumn{1}{c}{738} & 355 \\ 
\bottomrule
\end{tabular}
\label{tab:censoring_network_ratio}
\end{table}

\begin{table}[ht!]
\centering
\caption{Predictive performance for SurvivalGAN with and without the Censoring network. C-Index is better if higher and Brier score is better if lower. We see \methodname{} performs better without the censoring network.}
\begin{tabular}{c|cc|cc}
\toprule
\multirow{2}{*}{\textbf{Dataset}} & \multicolumn{2}{c|}{\textbf{SurvivalGAN}} & \multicolumn{2}{c}{\textbf{Censoring network}} \\ 
 & \multicolumn{1}{c}{\textbf{C-Index}} & \textbf{Brier score} & \multicolumn{1}{c}{\textbf{C-Index}} & \textbf{Brier score} \\
 \midrule
AIDS & \multicolumn{1}{c}{0.723$\pm$0.020} & \textbf{0.066$\pm$0.010} & \multicolumn{1}{c}{\textbf{0.730$\pm$0.010}} & 0.067$\pm$0.001 \\ 
CUTRACT & \multicolumn{1}{c}{\textbf{0.804$\pm$0.010}} & \textbf{0.083$\pm$0.010} & \multicolumn{1}{c}{0.770$\pm$0.010} & 0.100$\pm$0.001 \\ 
PHEART & \multicolumn{1}{c}{\textbf{0.644$\pm$0.010}} & \textbf{0.176$\pm$0.010} & \multicolumn{1}{c}{0.640$\pm$0.005} & 0.179$\pm$0.004 \\
SEER & \multicolumn{1}{c}{\textbf{0.834$\pm$0.010}} & \textbf{0.022$\pm$0.010} & \multicolumn{1}{c}{0.774$\pm$0.001} & 0.032$\pm$0.002 \\
METABRIC & \multicolumn{1}{c}{\textbf{0.719$\pm$0.020}} & \textbf{0.200$\pm$0.002} & \multicolumn{1}{c}{0.710$\pm$0.005} & 0.210$\pm$0.005 \\ 
\bottomrule
\end{tabular}
\label{tab:censoring_network_performance}
\end{table}

\subsection{SurvivalGAN vs. SurvivalVAE}

Our method can be adapted to other architectures as well to generate covariates. In this section, we perform another ablation by analyzing the performance of the synthetic data when using a variational autoencoder (\textbf{SurvivalVAE}) instead of a GAN.  For the experiment, we keep the same additional mechanisms in-place: imbalanced sampling around time and censoring, and the time-to-event/censoring regression. Table \ref{tab:predictive_perf_vae} contains the predictive performance of models trained on the synthetic data. We see that SurvivalGAN outperforms SurvivalVAE on the majority of datasets.

\clearpage

\begin{table}[ht!]
\centering
\caption{Predictive performance for models trained with synthetic data from SurvivalGAN vs. SurvivalVAE. C-Index is better if higher and Brier score is better if lower. SurvivalGAN tends to generate better quality data.}
\begin{tabular}{l|c|c|c}
\toprule
\textbf{Dataset} & \textbf{Method} & \textbf{C-Index} & \textbf{Brier score} 
\\ 
\midrule
\multirow{2}{*}{AIDS} 
& SurvivalVAE 
& \multicolumn{1}{c|}{0.638$\pm$0.020} & 0.058$\pm$0.000 
\\
 & SurvivalGAN 
 & \multicolumn{1}{c|}{\textbf{0.678$\pm$0.030}} & \textbf{0.057$\pm$0.010} 
 \\ 
\midrule
\multirow{2}{*}{CUTRACT} 
& SurvivalVAE 
 & \multicolumn{1}{c|}{0.791$\pm$0.010} & 0.103$\pm$0.003 
\\  
 & SurvivalGAN 
  & \multicolumn{1}{c|}{\textbf{0.799$\pm$0.020}} & \textbf{0.084$\pm$0.010} 
 \\
\midrule
\multirow{2}{*}{PHEART} 
& SurvivalVAE 
 & \multicolumn{1}{c|}{0.600$\pm$0.001} & 0.206$\pm$0.002
\\ 
 & SurvivalGAN 
 & \multicolumn{1}{c|}{\textbf{0.638$\pm$0.010}} & \textbf{0.181$\pm$0.010} 
 \\
  \midrule
\multirow{2}{*}{SEER} 
& SurvivalVAE 
& \multicolumn{1}{c|}{0.609$\pm$0.010} & 0.024$\pm$0.010
\\ 
 & SurvivalGAN
 & \multicolumn{1}{c|}{\textbf{0.835$\pm$0.010}} & \textbf{0.023$\pm$0.010} 
 \\
 \midrule
\multirow{2}{*}{METABRIC} 
& SurvivalVAE 
& \multicolumn{1}{c|}{0.724$\pm$0.010} & 0.191$\pm$0.002
\\ 
 & SurvivalGAN 
 & \multicolumn{1}{c|}{\textbf{0.734$\pm$0.010}} & \textbf{0.189$\pm$0.010}
 \\
\bottomrule
\end{tabular}
\label{tab:predictive_perf_vae}
\end{table}

\subsection{SurvivalGAN vs. SurvivalGMM}

We test \methodname{} against the simplest possible generative model for the covariates, a Gaussian Mixture Model (SurvivalGMM). This uses the same Gaussian Mixture Model that is used in the tabular encoder and class encoder with 100 mixture components (see Section \ref{subsection:encoders} for information on these), but now to generate covariates instead of using the GAN. We provide the downstream performances in Table \ref{tab:predictive_perf_gmm}. We see that using a GAN performs better than using the GMM.

\begin{table}[H]
\centering
\caption{Predictive performance for models trained with synthetic data from SurvivalGAN vs. SurvivalGMM. C-Index is better if higher and Brier score is better if lower. We see SurvivalGAN is significantly better than SurvivalGMM.}
\begin{tabular}{c|c|cc}
\toprule
\textbf{Dataset} & \textbf{Method} & \textbf{C-Index} & \textbf{Brier score} 
 \\
\midrule
\multicolumn{1}{l|}{\multirow{2}{*}{AIDS}}
& SurvivalGMM 
 & \multicolumn{1}{c|}{0.510$\pm$0.157} & 0.061$\pm$0.002 
\\ 
\multicolumn{1}{l|}{} & SurvivalGAN
 & \multicolumn{1}{c|}{\textbf{0.678$\pm$0.030}} & \textbf{0.057$\pm$0.010} 
\\
\midrule
\multicolumn{1}{l|}{\multirow{2}{*}{CUTRACT}}
& SurvivalGMM 
& \multicolumn{1}{c|}{0.780$\pm$0.008} & 0.089$\pm$0.001 
\\ 
\multicolumn{1}{l|}{} 
& SurvivalGAN 
 & \multicolumn{1}{c|}{\textbf{0.799$\pm$0.020}} & \textbf{0.084$\pm$0.010} 
\\
\midrule
\multicolumn{1}{l|}{\multirow{2}{*}{PHEART}} 
& SurvivalGMM 
 & \multicolumn{1}{c|}{0.627$\pm$0.002} & 0.205$\pm$0.001 
\\ 
\multicolumn{1}{l|}{} 
& SurvivalGAN 
 & \multicolumn{1}{c|}{\textbf{0.638$\pm$0.010}} & \textbf{0.181$\pm$0.010} 
\\
\midrule
\multicolumn{1}{l|}{\multirow{2}{*}{SEER}}
& SurvivalGMM 
 & \multicolumn{1}{c|}{0.662$\pm$0.017} & 0.024$\pm$0.010 
\\ 
\multicolumn{1}{l|}{} 
& SurvivalGAN 
 & \multicolumn{1}{c|}{\textbf{0.835$\pm$0.010}} & \textbf{0.023$\pm$0.010} 
\\
\midrule
\multicolumn{1}{l|}{\multirow{2}{*}{METABRIC}}
& SurvivalGMM 
 & \multicolumn{1}{c|}{0.564$\pm$0.013} & 0.282$\pm$0.036 
\\
\multicolumn{1}{l|}{}
& SurvivalGAN 
 & \multicolumn{1}{c|}{\textbf{0.734$\pm$0.010}} & \textbf{0.189$\pm$0.010}
\\ 
\bottomrule
\end{tabular}
\label{tab:predictive_perf_gmm}
\end{table}

\clearpage
\section{QUALITATIVE RESULTS}
\label{app:qualitative}

Here we provide qualitative results in the form of t-SNE plots for the covariates \citep{van2008visualizing} and Kaplan-Meier plots for the time and event \citep{kaplan1958nonparametric}.

\subsection{Downstream Predictive Models}

\begin{figure}[ht!]
\includegraphics[width=\textwidth]{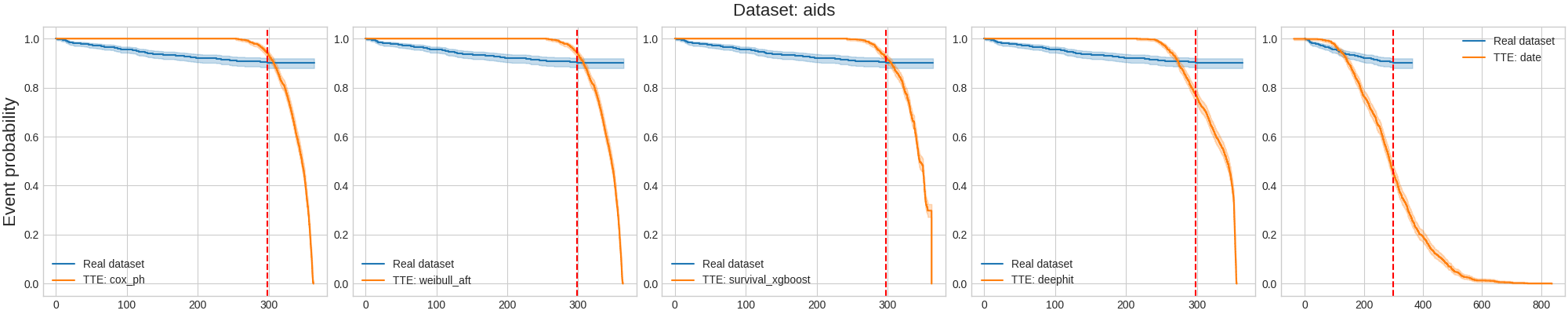}
\includegraphics[width=\textwidth]{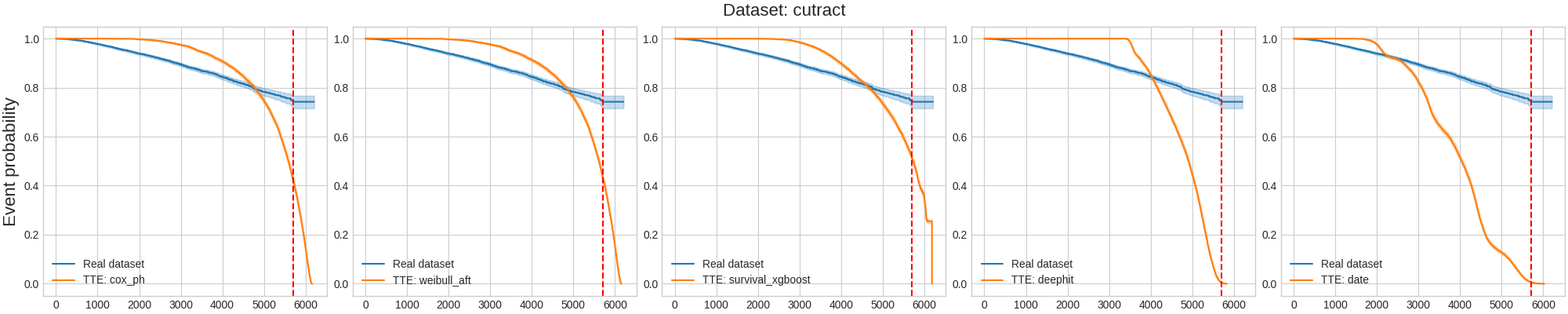}
\includegraphics[width=\textwidth]{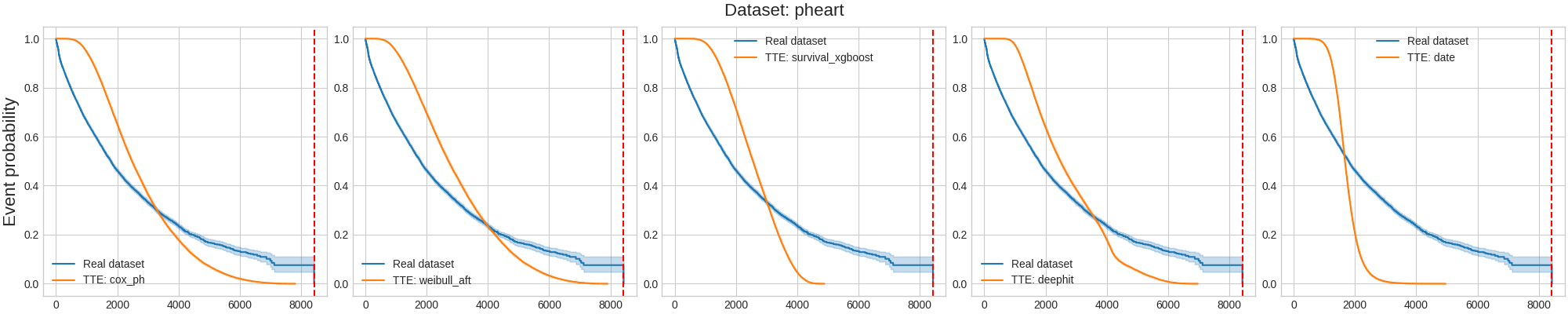}
\includegraphics[width=\textwidth]{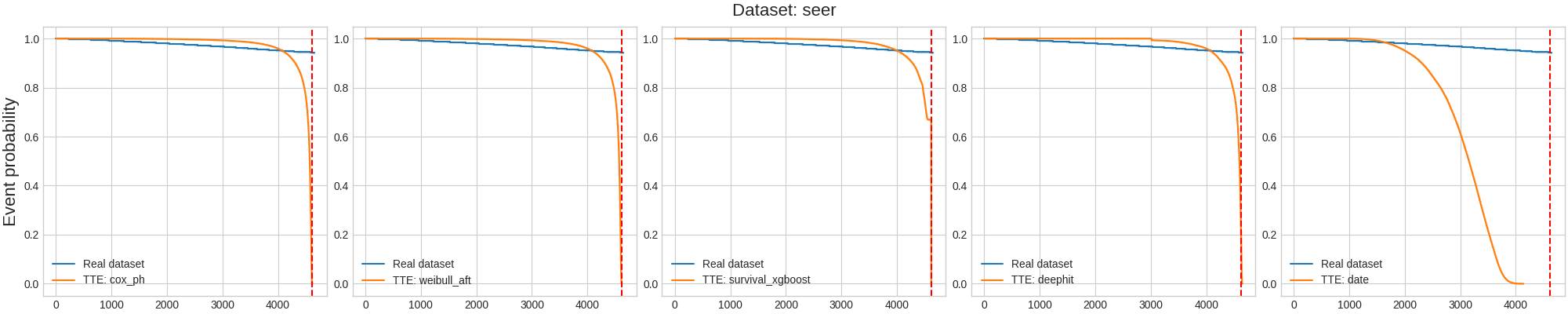}
\caption{Kaplan-Meier plots of reference time-to-event models. The first five survival-function-based methods tend to be over-optimistic, while the last model is over-pessimistic.}
\label{fig:km_plot_tte_models_appendix}
\end{figure}

Figure \ref{fig:km_plot_tte_models_appendix} presents the Kaplan-Meier plots for the time-to-event models (the downstream models). The observed trend is visible in the datasets, leading to over-optimistic or over-pessimistic time-to-event values, and it supports the need for a reliable method to overcome the time-to-event/censoring problem.

\clearpage
\subsection{Data Fidelity and Diversity}
\label{app:tsne_km}

\begin{figure}[ht!]
\centering
\includegraphics[width=0.8\textwidth]{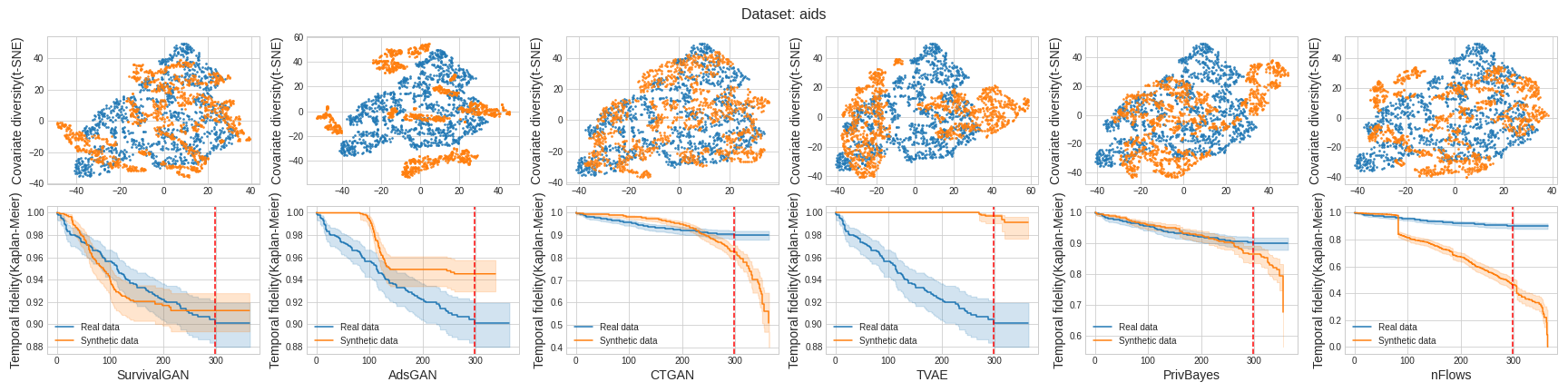}
\includegraphics[width=0.8\textwidth]{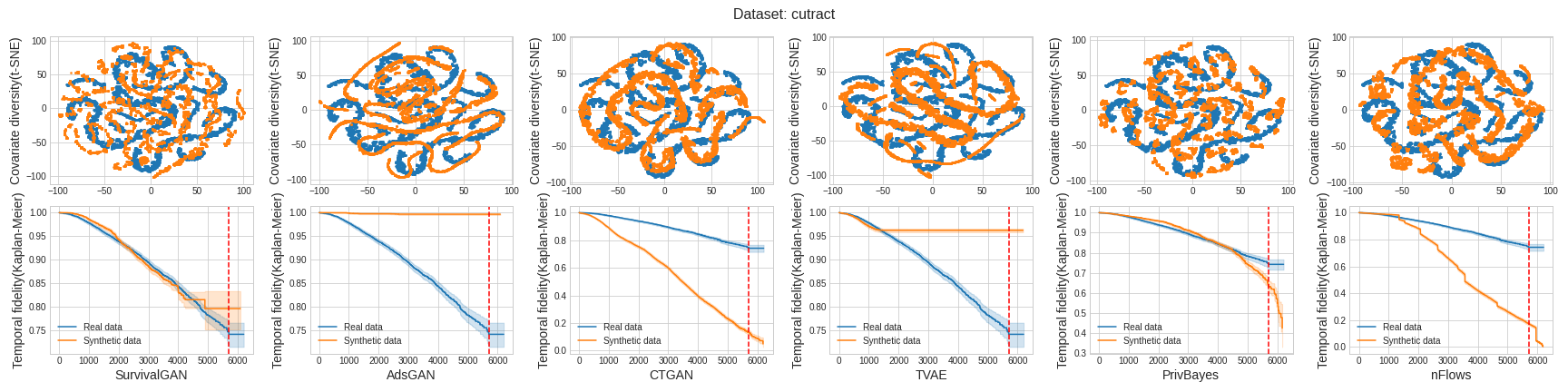}
\includegraphics[width=0.8\textwidth]{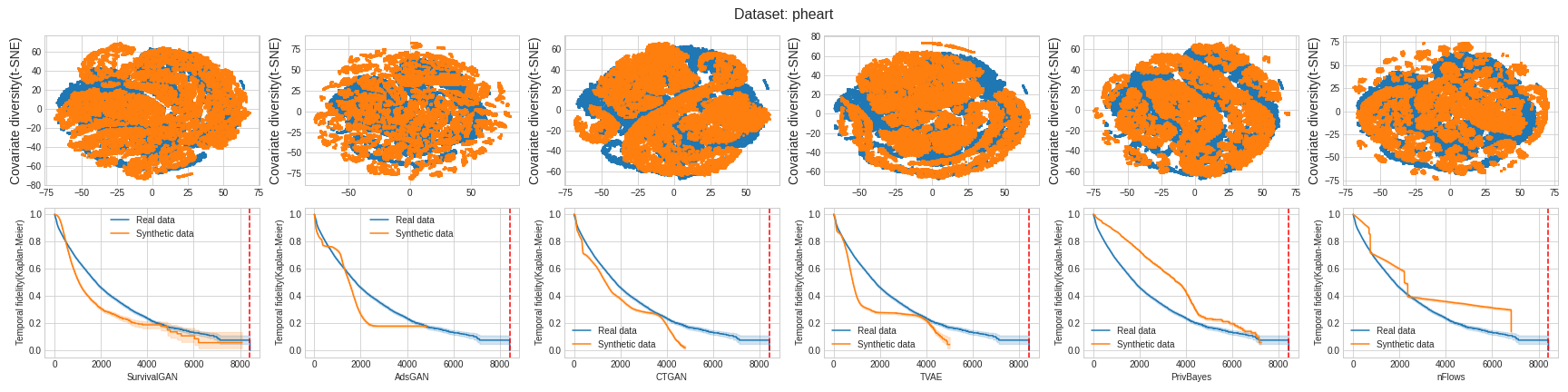}
\includegraphics[width=0.8\textwidth]{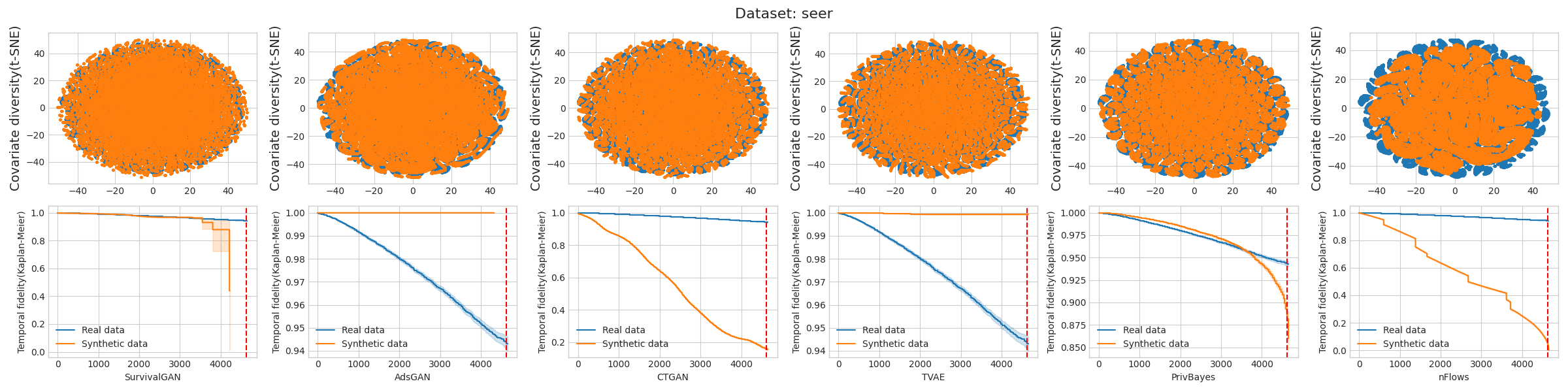}
\includegraphics[width=0.8\textwidth]{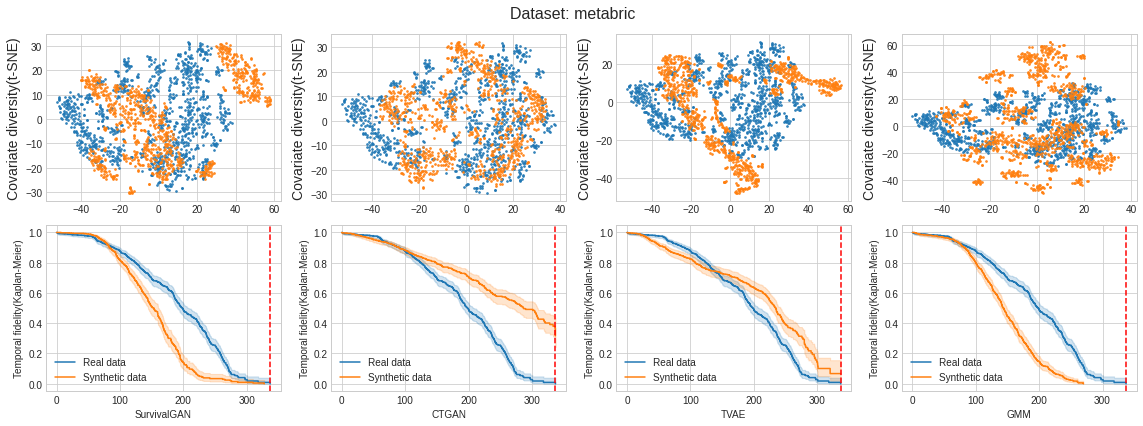}
\caption{
Data diversity visualization for all datasets, from a single random seed. For each dataset, the 1st row contains the t-SNE plots on the covariates, and the 2nd row contains the Kaplan-Meier plots for time and censoring. Each column provides the visualization for each of the available benchmarks.}
\label{fig:experiments_fidelity_plots_appendix}
\end{figure}

Figure \ref{fig:experiments_fidelity_plots_appendix}, includes the t-SNE plots for covariate coverage and the Kaplan-Meier visualizations for temporal fidelity, using all datasets. Qualitatively, we see SurvivalGAN is robust in generating the covariate and temporal distributions across all datasets. The t-SNE plots show the covariates typically cover the data distribution at least as well as the other baselines. More importantly, the KM plots show the ability to model $p(t, E)$ is significantly better for \methodname{} than the baselines.


\clearpage

\subsection{Sources of Gain}
\label{app:sources_of_gain}

Figure \ref{fig:experiments_gains_km} reports the sources of gain visualizations for all datasets (apart from METABRIC) with t-SNE for the covariates and KM plots for the time/censoring. We observe that the conditional GAN has an important impact on the quality of the covariates (final column of the t-SNE plots), while all the components contribute to the temporal calibration and sightedness (KM plots).

\begin{figure}[h]
\centering
\includegraphics[width=0.75\textwidth]{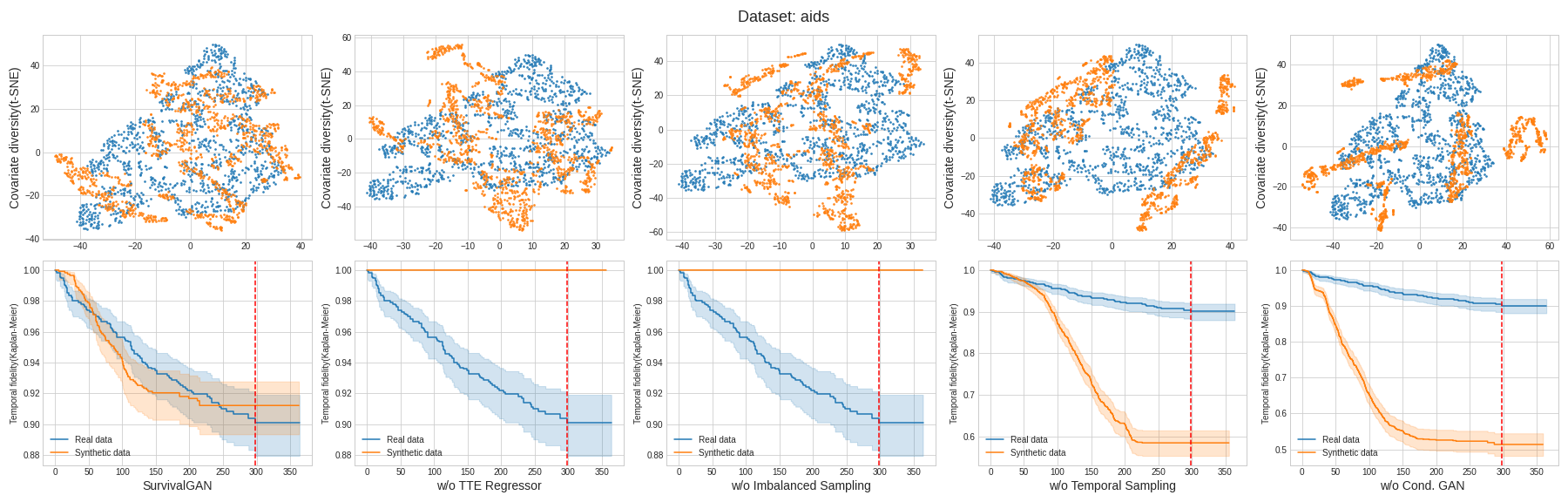}
\includegraphics[width=0.75\textwidth]{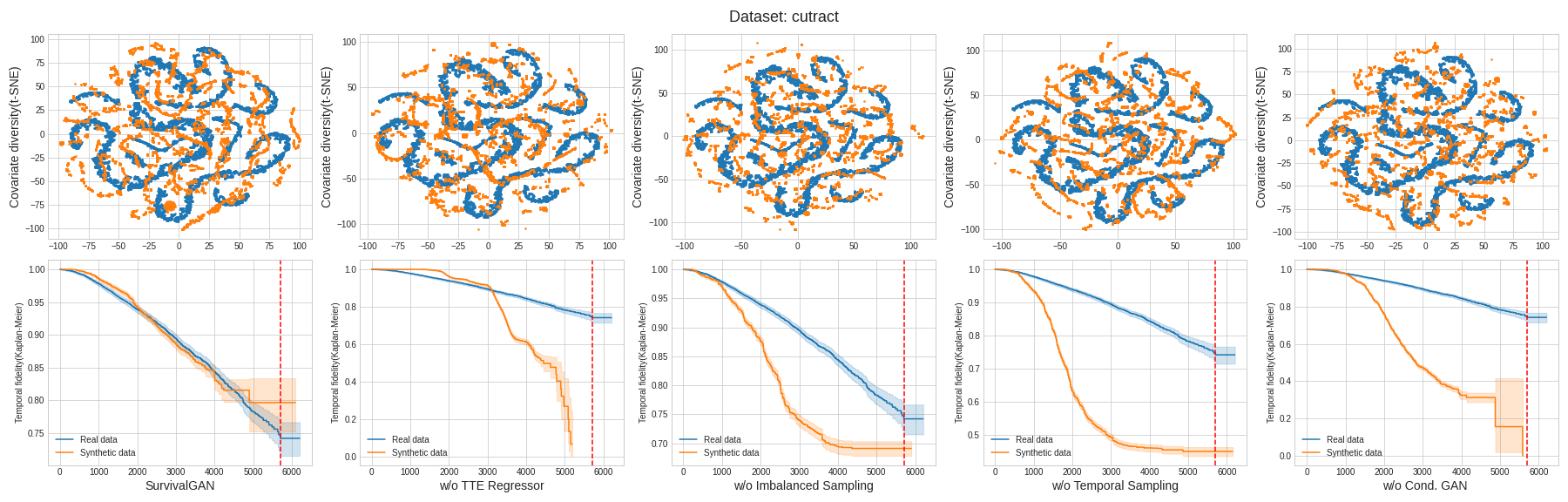}
\includegraphics[width=0.75\textwidth]{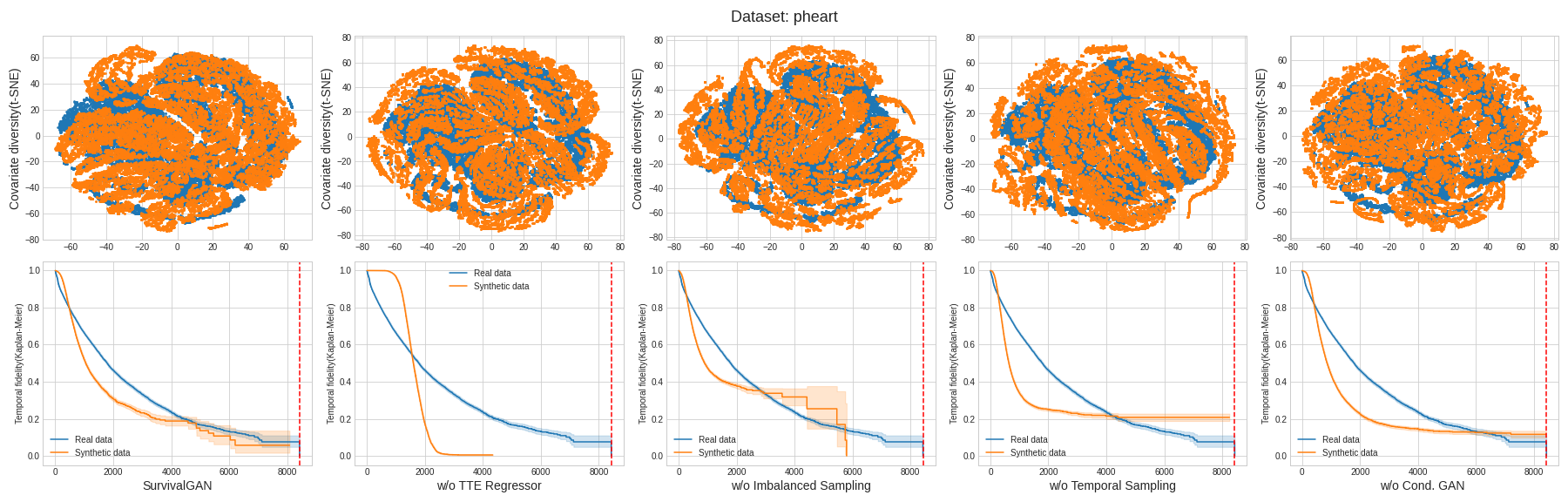}
\includegraphics[width=0.75\textwidth]{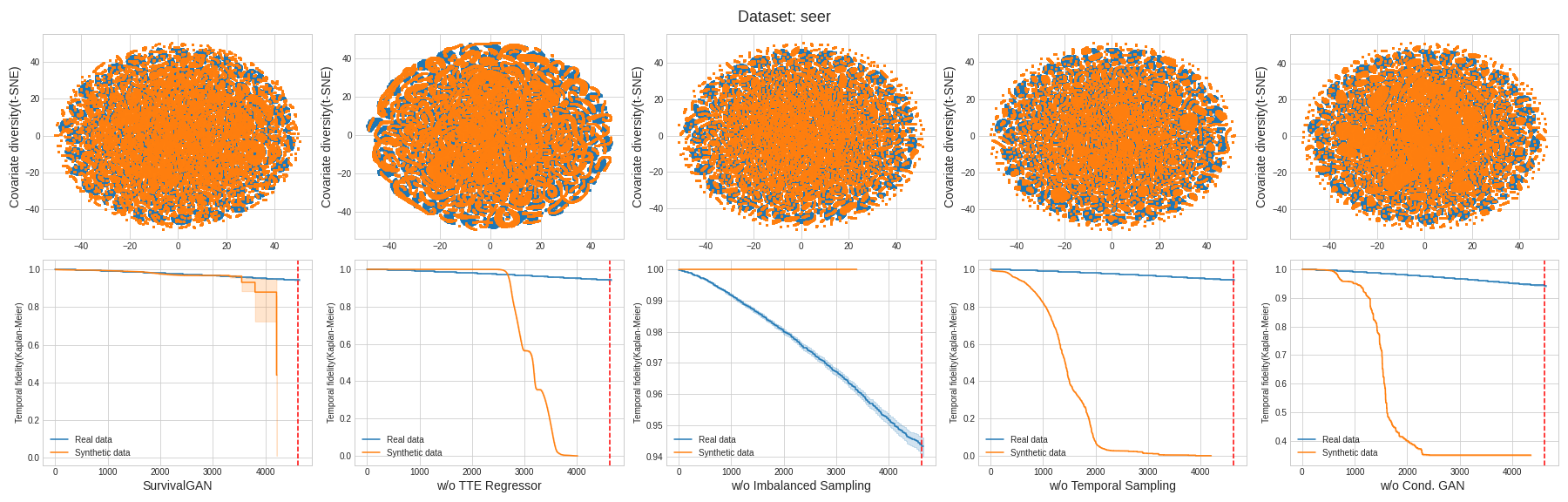}
\caption{Sources of gain visualizations using t-SNE and Kaplan-Meier plots. For each dataset, the 1st row contains the t-SNE plots of the covariates, and the 2nd row contains the Kaplan-Meier plots for time and censoring. Each column corresponds to a given ablation scenario (certain component missing).}
\label{fig:experiments_gains_appendix}
\end{figure}

\clearpage

\subsection{SurvivalGAN vs. SurvivalVAE}

\begin{figure}[H]
\includegraphics[width=\textwidth]{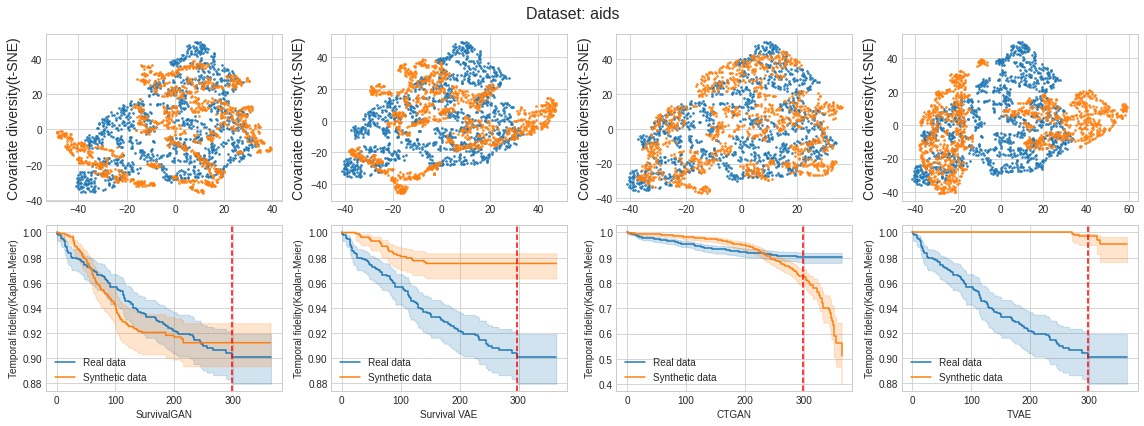}
\includegraphics[width=\textwidth]{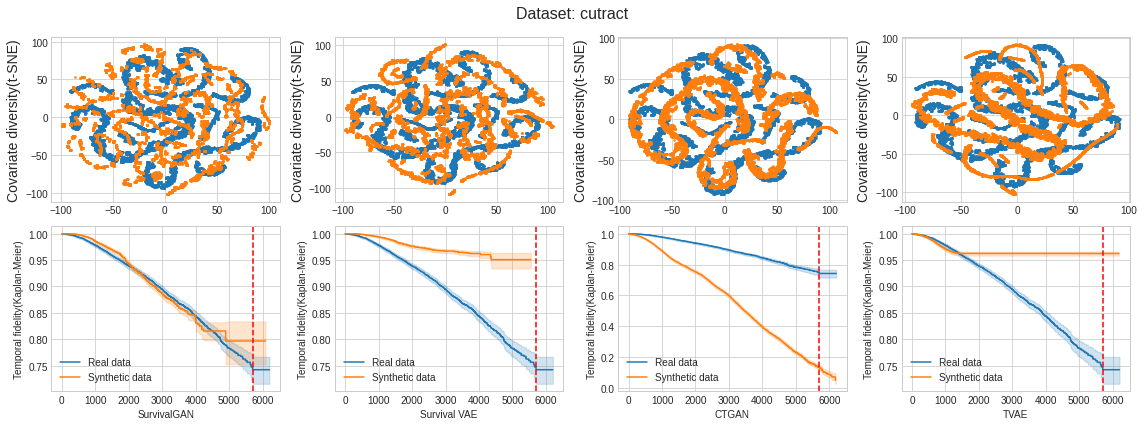}
\includegraphics[width=\textwidth]{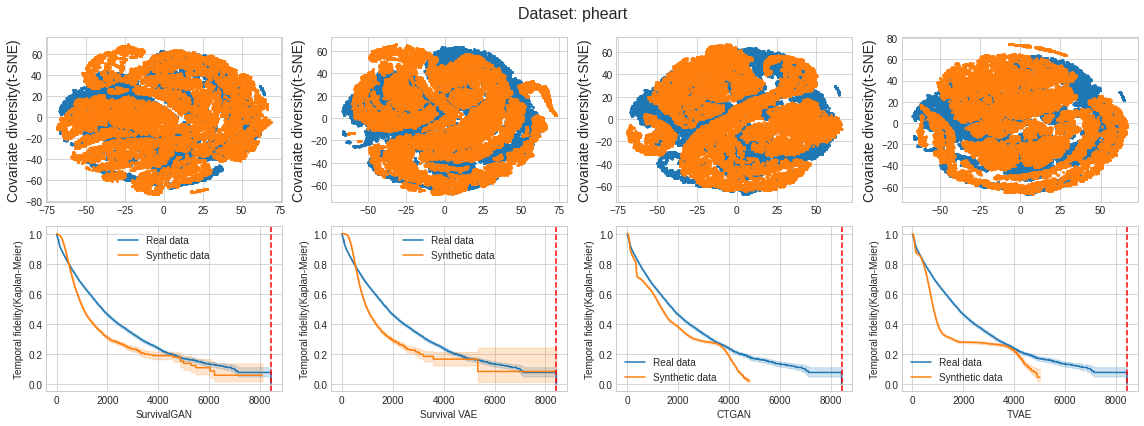}
\caption{
Data diversity visualization for SurvivalVAE, from a single random seed. For each dataset, the 1st row contains the t-SNE plots of the covariates, and the 2nd row contains the Kaplan-Meier plots for time and censoring. Each column provides the visualization for each of the 4 benchmarks.}
\label{fig:experiments_fidelity_plots_vae}
\end{figure}

Finally, Figure \ref{fig:experiments_fidelity_plots_vae} shows the qualitative differences between \methodname{} and SurvivalVAE. The biggest difference is in the Kaplan-Meier plots, showing \methodname{} models $p(t, E)$ more faithfully than SurvivalVAE. We include CTGAN and TVAE as well to show that generally VAE based models are more over-optimistic than GAN based models.

\vfill

\end{document}